%% file: main.tex
\documentclass[10pt,twocolumn,letterpaper]{article}

\usepackage[pagenumbers]{cvpr} 
\usepackage{amsfonts, amsmath, amssymb, amsthm}
\usepackage{multirow}  
\usepackage{numprint}
\usepackage{booktabs}  
\usepackage{comment}
\usepackage[normalem]{ulem}  
\usepackage[inline]{enumitem}

\definecolor{cvprblue}{rgb}{0.21,0.49,0.74}
\usepackage[pagebackref,breaklinks,colorlinks,citecolor=cvprblue]{hyperref}
\usepackage[capitalize]{cleveref}

\def\paperID{11521} 
\def\confName{CVPR}
\def\confYear{2024}
\input{fancy-math-fonts}

\input{abbreviations}

\input{custom-commands}

\def\Title{1-Lipschitz Layers Compared: Memory, Speed, and Certifiable Robustness}
\title{\Title}
\input{authors}
\begin{document}
\maketitle
\footnotetext[0]{\textasteriskcentered Joined first authors.}

\input{sec/0_abstract}

\input{sec/1_introduction}

\input{sec/2_LipschitzLayers}

\input{sec/4_comparison}

\input{sec/5a_experiments}
\input{sec/5b_results}
\input{sec/7_Conclusion}

{\small
\bibliographystyle{ieee_fullname}
\bibliography{biblio}
}
\onecolumn
\appendix 
\input{sec/9_appendix}
\pagebreak

\end{document}

%% file: fancy-math-fonts.tex
\newcommand{\R}{\mathbb{R}}

\newcommand{\MM}{\mathcal{M}}

\newcommand{\OO}{\mathcal{O}}

\newcommand{\xx}{\mathbf{x}}

%% file: abbreviations.tex
\newcommand{\stdconv}{Standard\xspace}
\newcommand{\powermeth}{power method\xspace}

\newcommand{\AOL}{AOL\xspace}
\newcommand{\BnB}{Bjorck \& Bowie\xspace}
\newcommand{\Cayley}{Cayley\xspace}
\newcommand{\CPL}{CPL\xspace}
\newcommand{\SOC}{SOC\xspace}
\newcommand{\ECO}{ECO\xspace}
\newcommand{\SLL}{SLL\xspace}
\newcommand{\LOT}{LOT\xspace}
\newcommand{\BCOP}{BCOP\xspace}
\newcommand{\Sandwich}{Sandwich\xspace}

\newcommand{\MaxMin}{MaxMin\xspace}

\newcommand{\CIFARS}{CIFAR-10\xspace}
\newcommand{\CIFARL}{CIFAR-100\xspace}
\newcommand{\TinyIN}{Tiny ImageNet\xspace}

\newcommand{\MACS}{MACS}

\newcommand{\Inference}{Inference\xspace}  
\newcommand{\inference}{inference\xspace}  

\newcommand{\Cra}{Certified robust accuracy\xspace}
\newcommand{\cra}{certified robust accuracy\xspace}
\newcommand{\ols}{$1$-Lipschitz\xspace}
\newcommand{\lsc}{Lipschitz constant\xspace}
\newcommand{\fft}{fast Fourier transform \xspace}
\newcommand{\Ord}[1]{\mathcal{O}(#1)}

\newcommand{\CONV}{CONV\xspace}
\newcommand{\FFT}{FFT\xspace}
\newcommand{\MATMUL}{MM\xspace}
\newcommand{\MV}{MV\xspace}  
\newcommand{\ACT}{ACT\xspace}  
\newcommand{\INV}{INV\xspace}

\newcommand{\ONI}{ONI\xspace}

\renewcommand{\vec}[1]{\textbf{#1}}

\newcommand{\bp}[1]{{\footnotesize\textcolor{blue}{[BP: #1]}}}
\newcommand{\fb}[1]{{\footnotesize\textcolor{orange}{[FB: #1]}}}

\newcommand{\spect}[1]{\|#1\|_\text{2}}
\newcommand{\Reals}{\mathbb{R}}

\newcommand{\mrtimes}[1]{\multirow{2}{*}{$#1\times \Big\{$}}

%% file: custom-commands.tex
\theoremstyle{definition}

\newtheorem{remark}{Remark}

\newcommand{\ReLU}[1]{\mathrm{ReLU}\left(#1\right)}
\newcommand{\diag}[1]{\mathrm{diag}\left(#1\right)}


%% file: authors.tex
\author{
    Bernd Prach,\negmedspace\textsuperscript{1,\textasteriskcentered} \
    Fabio Brau,\negmedspace\textsuperscript{2,\textasteriskcentered} \ 
    Giorgio Buttazzo,\negmedspace\textsuperscript{2} \
    Christoph H. Lampert\textsuperscript{1}
    \\
    \textsuperscript{1} ISTA, Klosterneuburg, Austria \\
    \textsuperscript{2} Scuola Superiore Sant'Anna, Pisa, Italy \\
    {\tt\small 
        \{bprach, chl\}@ist.ac.at,
        \{fabio.brau, giorgio.buttazzo\}@santannapisa.it
    }
}


%% file: sec/0_abstract.tex
\begin{abstract}
    The robustness of neural networks against input perturbations with bounded magnitude represents a serious concern in the deployment of deep learning models in safety-critical systems. Recently, the scientific community has focused on enhancing certifiable robustness guarantees by crafting \ols neural networks that leverage Lipschitz bounded dense and convolutional layers. Although different methods have been proposed in the literature to achieve this goal, understanding the performance of such methods is not straightforward, since different metrics can be relevant (e.g., training time, memory usage, accuracy, certifiable robustness) for different applications. For this reason, this work provides a thorough theoretical and empirical comparison between methods by evaluating them in terms of memory usage, speed, and certifiable robust accuracy. The paper also provides some guidelines and recommendations to support the user in selecting the methods that work best depending on the available resources.
    We provide code at 
    {\footnotesize \tt
    \href{https://github.com/berndprach/1LipschitzLayersCompared}{github.com/berndprach/1LipschitzLayersCompared}.
    }
\end{abstract}

%% file: sec/1_introduction.tex
\begin{figure}[t!]
\centering
\includegraphics[width=0.96\columnwidth]{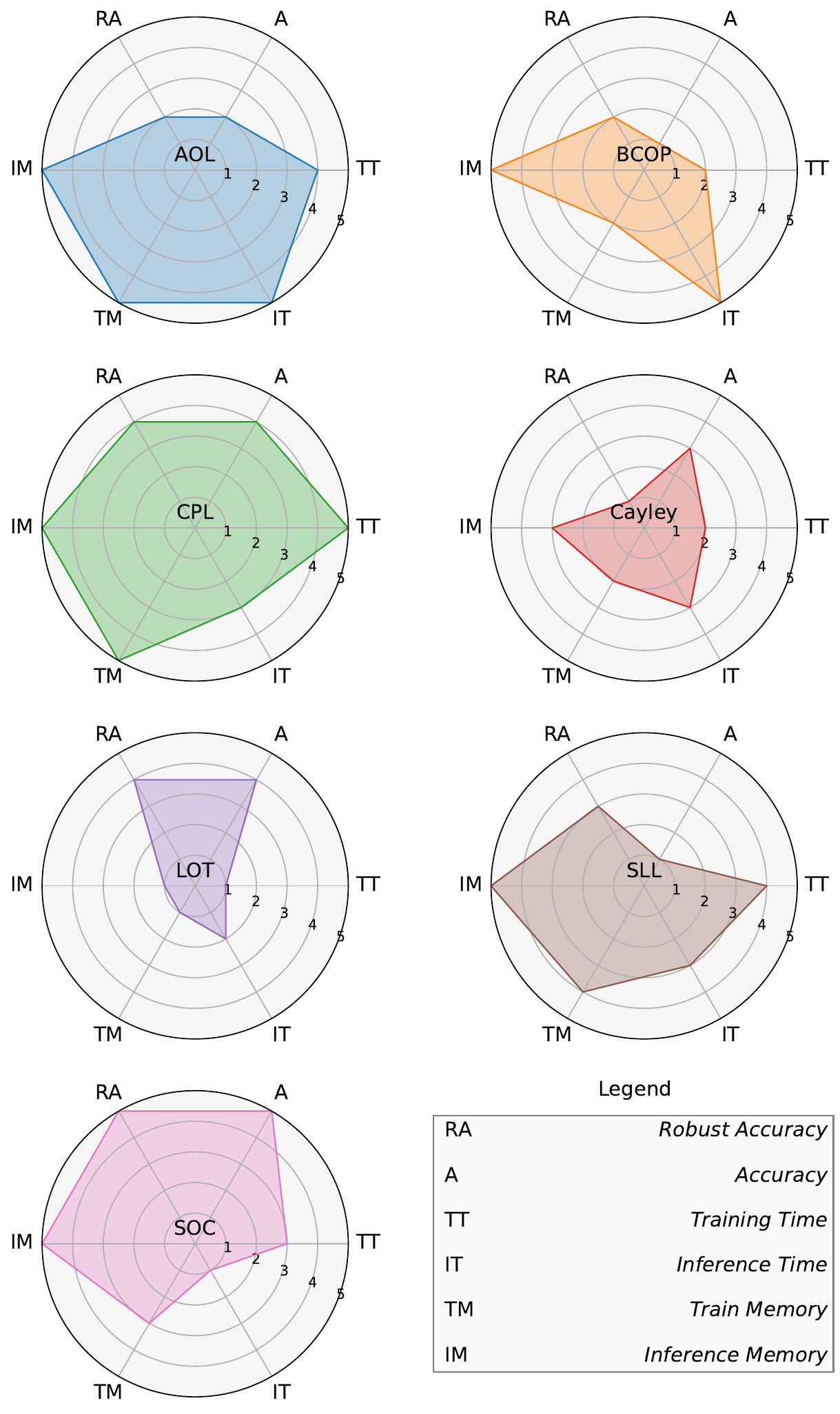}
\caption{
    Evaluation of \ols methods on different metrics. Scores are assigned from 1 (worst) to 5 (best) to every method based on the results reported in 
    \Cref{sec:theoretical,sec:results}.
} 
\label{fig:radar_plot}
\end{figure}

\section{Introduction}

Modern artificial neural networks achieve high accuracy and sometimes superhuman performance in many different tasks, but it is widely recognized that they are not robust to tiny and imperceptible input perturbations \cite{szegedy, biggio} that, if properly crafted, can cause a model to produce the wrong output. Such inputs, known as \textit{Adversarial Examples}, represent a serious concern for the deployment of machine learning models in safety-critical systems \cite{li2023sok}. For this reason, the scientific community is pushing towards \textit{guarantees} of robustness. Roughly speaking, a model $f$ is said to be $\varepsilon$-robust for a given input $x$ if no perturbation of magnitude bounded by $\varepsilon$ can change its prediction. Recently, in the context of image classification, various approaches have been proposed to achieve certifiable robustness, including \textit{Verification, Randomized Smoothing}, and \textit{Lipschitz bounded Neural Networks}.
\medskip

Verification strategies aim to establish, for any given model, whether all samples contained in a $l_2$-ball with radius $\varepsilon$ and centered in the tested input $x$ are classified with the same class as $x$. In the exact formulation, verification strategies involve the solution of an NP-hard problem \cite{Katz_Kochenderfer_2017}. Nevertheless, even in a relaxed formulation, \cite{Wong_Kolter_2018}, these strategies 
require a huge computational effort \cite{Weng_Daniel_2018b}.
\medskip

Randomized smoothing strategies, initially presented in \cite{Cohen_Kolter_2019}, represent an effective way of crafting a certifiable-robust classifier $g$ based on a base classifier $f$. If combined with an additional denoising step, they can achieve state-of-the-art levels of robustness, \cite{carlini2023certified}. However, since they require multiple evaluations of the base model (up to 100k evaluations) for the classification of a single input, they cannot be used for real-time applications.
\medskip

Finally, Lipschitz Bounded Neural Networks represent a valid alternative to produce certifiable classifiers, since they only require a single forward pass of the model at inference time \cite{losch2023certified,parseval,aol,bcop,cayley,GloRo,brau}. 
Indeed, the difference between the two largest output components of the model directly provides a lower-bound, in Euclidean norm, of the minimal adversarial perturbation capable of fooling the model. 
Lipschitz-bounded neural networks can be obtained by the composition of $1$-Lipschitz layers \cite{anil_fc}.
The process of parameterizing \ols layers
is fairly straightforward for fully connected layers.
However, for convolutions --- with overlapping kernels\ --- deducing an effective parameterization is a hard problem.
Indeed, the Lipschitz condition can be essentially thought of as a condition on the Jacobian of the layer.
However, the Jacobian matrix can not be efficiently computed.
\medskip

In order to avoid the explicit computation of the Jacobian, 
various methods have been proposed, including 
parameterizations that cause the Jacobian to be (very close to)
orthogonal \cite{soc,cayley,lot,bcop} 
and methods that rely on an upper bound on the Jacobian instead \cite{aol}.
Those different methods differ drastically in training and validation
requirements (in particular time and memory) as well as empirical
performance. Furthermore, increasing training time or model sizes
very often also increases the empirical performance.
This makes it hard to judge from the existing literature 
which methods are the most promising.
This becomes even worse when working with specific computation
requirements, such as restrictions on the available memory.
In this case, it is important to
choose the method that better suits the characteristics of the system in terms of evaluation time, memory usage as well and certifiable-robust-accuracy.
\medskip

\noindent\textbf{This works}
aims at giving a 
comprehensive comparison of different strategies for crafting \ols layers from both a theoretical and practical perspective.
For the sake of fairness, we consider several metrics such as
\textit{Time} and \textit{Memory} requirements for both training and inference, \textit{Accuracy}, as well as \textit{Certified Robust Accuracy}.
The main contributions are the following: \\
\begin{itemize}
    \item 
    An empirical comparison of 1-Lipschitz layers based on six different metrics, and three different datasets on four architecture sizes with three time constraints.
    \item A theoretical comparison of the runtime complexity and the memory usage of existing methods.
    \item A review of the most recent methods in the literature, including implementations with a revised code that we will release publicly for other researchers to build on. 
\end{itemize}

%% file: sec/2_LipschitzLayers.tex
\section{Existing Works and Background}
\label{sec:methods}
In recent years, various methods have been proposed
for creating artificial neural networks with a
bounded Lipschitz constant.
The \emph{\lsc} of a function $f:\Reals^n \rightarrow \Reals^m$ 
with respect to the $l_2$ norm
is the smallest $L$ such that for all $x,y \in \Reals^n$
\begin{align} \label{eq:Lipschitz}
    \|f(x) - f(y)\|_2 \le L \|x-y\|_2.
\end{align} We also extend this definition to networks and layers, by considering the $l_2$ norms of the flattened input and output tensors in \Cref{eq:Lipschitz}.
A layer is called \ols if its \lsc is at most 1.
For linear layers, the Lipschitz constant is equal
to the \emph{spectral norm} of the weight matrix that is given as
\begin{align}
    \spect{M} = \sup_{\vec{v}\ne 0} \frac{\|M\vec{v}\|_2}{\|\vec{v}\|_2}.
\end{align} A particular class of linear \ols layers are ones with an orthogonal Jacobian matrix.
The Jacobian matrix of a layer is the matrix of partial derivatives of the flattened outputs
with respect to the flattened inputs. 
A matrix $M$ is orthogonal if $MM^\top = I$, where $I$ is the identity matrix.
For layers with an orthogonal Jacobian, 
\Cref{eq:Lipschitz} always holds with equality and, because of this, a lot of methods aim at constructing such \ols layers.
\medskip

All the neural networks analyzed in this paper consist of \ols parameterized layers and \ols activation functions, with no skip connections and no batch normalization. Even though the commonly used ReLU activation function is $1$-Lipschitz, Anil \etal \cite{anil_fc} showed that it reduces the expressive capability of the model. Hence, we adopt the \MaxMin activation  proposed by the authors and commonly used in \ols models.
Concatenations of \ols functions are \ols, so the
networks analyzed are \ols by construction.

\subsection{Parameterized \ols Layers}
This section provides an overview of the existing methods for providing \ols layers. 
We discuss fundamental methods and for estimating the spectral norms of linear and convolutional layers, i.e. \textit{Power Method}~\cite{miyato} and \textit{Fantistic4}~\cite{fantastic4}, and for crafting orthogonal matrices, i.e. \BnB~\cite{Bjorck_1971_SIAM}, in \Cref{sec:bounding_methods}. 
The rest of this section describes 7 methods from the literature that construct \ols convolutions: \BCOP, \Cayley, \SOC, \AOL, \LOT, \CPL, and \SLL.
Further \ols methods, \cite{eco, sandwich_2023_wang, ONI_2020_huang}, and the reasons why they were not included in our main comparison can be found in \Cref{sec:omitted}.
\paragraph{\BCOP}
\emph{Block Orthogonal Convolution Parameterization (\BCOP)} was introduced by Li \etal in \cite{bcop} to extend a previous work by Xiao \etal~\cite{dynamical_isometry} that focused on the importance of orthogonal initialization of the weights.
For a $k \times k$ convolution, 
\BCOP uses a set of $(2k-1)$ parameter matrices.
Each of these matrices is orthogonalized using the algorithm by \BnB \cite{Bjorck_1971_SIAM} (see also \Cref{sec:bnb}).
Then, a $k \times k$ kernel is constructed
from those matrices to guarantee that the resulting layer is orthogonal.

\paragraph{Cayley}
Another family of orthogonal convolutional and fully connected layers has been proposed by Trockman and Kolter \cite{cayley} by leveraging the \textit{Cayley Transform} \cite{Cayley_1846_JRAM}, which maps a skew-symmetric matrix $A$ into an orthogonal matrix $Q$ using the relation
\begin{equation}
\label{eq:cayley}
 Q=(I-A)(I+A)^{-1}.
\end{equation}
The transformation can be used to parameterize
orthogonal weight matrices for linear layers
in a straightforward way.
For convolutions, the authors make use of the fact
that circular padded convolutions
are vector-matrix products in the Fourier domain.
As long as all those vector-matrix products have orthogonal matrices, the full convolution will have an orthogonal Jacobian.
For \emph{Cayley Convolutions}, those matrices are orthogonalized using the Cayley transform.

\paragraph{SOC}
\textit{Skew Orthogonal Convolution} is an orthogonal convolutional layer presented by Singla \etal~\cite{soc}, obtained by leveraging the exponential convolution \cite{Hoogeboom}.
Analogously to the matrix case, given a kernel
$L\in\R^{c\times c \times k\times k}$, 
the exponential convolution can be defined as
\begin{equation}
    \exp(L)(x):=x + \frac{L \star x}{1} + \frac{L\star^2 x}{2!} + \cdots + \frac{L\star^k x}{k!}+\cdots,
\end{equation}
where $\star^k$ denotes a convolution applied $k$-times.
The authors proved that any exponential convolution has an orthogonal Jacobian matrix as long as $L$ is skew-symmetric, providing a way of parameterizing \ols layers.
In their work, the sum of the infinite series is approximated by computing only the first $5$ terms during training and the first $12$ terms during the inference, and $L$ is normalized to have unitary spectral norm following the method presented in \cite{fantastic4} (see \Cref{sec:fantastic_four}).

\paragraph{AOL}
Prach and Lampert \cite{aol} introduced 
\emph{Almost Orthogonal Lipschitz (AOL)} layers.
For any matrix $P$, they defined a diagonal rescaling matrix $D$ with 
\begin{align} \label{eq:aol}
        D_{ii} = \Big( \sum_j \left|P^\top P \right|_{ij} \Big)^{-1/2}
\end{align}
and proved that the spectral norm of $PD$ is bounded by 1.
This result was used to show that the linear layer given by
$l(x) = PDx + b$
(where $P$ is the learnable matrix and $D$ is given by \cref{eq:aol}) is \ols.
Furthermore, the authors extended the idea
so that it can also be efficiently applied to convolutions.
This is done by calculating the rescaling in \Cref{eq:aol}
with the Jacobian $J$ of a convolution instead of $P$. 
In order to evaluate it efficiently the authors express the
elements of $J^\top J$ explicitly in terms of the kernel values.

\paragraph{LOT}
The layer presented by Xu \etal~\cite{lot} extends the idea
of \cite{ONI_2020_huang} to use the \textit{Inverse Square Root} of a matrix
in order to orthogonalize it.
Indeed, for any matrix $V$, 
the matrix $Q=V(V^TV)^{-\frac12}$
is orthogonal. 
Similarly to the \textit{Cayley} method, 
for the \emph{layer-wise orthogonal training (LOT)}
the convolution is applied in the Fourier frequency domain.
To find the inverse square root, the authors relay on an iterative \textit{Newton Method}. In details, defining  $Y_0=V^TV$, $Z_0=I$, and
\begin{equation}
    Y_{i+1} = \frac12 Y_i\left(3I-Z_iY_i\right),\, Z_{i+1} = \frac12\left(3I-Z_iY_i\right)Z_i,
\end{equation}
it can be shown that $Y_i$ converges to $(V^TV)^{-\frac12}$. 
In their proposed layer, the authors apply $10$ iterations of the method for both training and evaluation.

\paragraph{CPL}
Meunier \etal~\cite{CPL} proposed the \emph{Convex Potential Layer}.
Given a non-decreasing \ols function $\sigma$ (usually ReLU), the layer is constructed as
\begin{equation} \label{eq:cpl}
    l(x) = x - \frac{2}{\|W\|_2^2} W^\top \sigma(Wx + b),
\end{equation}
which is \ols by design.
The spectral norm required to calculate $l(x)$ is approximated using 
the \powermeth (see \Cref{sec:power_method}).

\paragraph{SLL}
The \emph{SDP-based Lipschitz Layers (SLL)} proposed by Araujo \etal~\cite{SLL} combine the \CPL layer
with the upper bound on the spectral norm from AOL.
The layer can be written as 
\begin{align} \label{eq:sll}
    l(x) = x - 2W Q^{-2} D^2 \sigma \left(W^\top x + b\right),
\end{align}
where $Q$ is a learnable diagonal matrix with positive entries and $D$ is deduced by applying \Cref{eq:aol} to $P=WQ^{-1}$.

\begin{remark}
Both \CPL and \SLL are non-linear by construction, so they can be used to construct a network without any further use of activation functions.
However, carrying out some preliminary experiments, we empirically found
that alternating \CPL (and \SLL) layers
with \MaxMin activation layers allows achieving a better performance.
\end{remark}
\input{tab/theoretical_new}

%% file: tab/theoretical_new.tex
\newcommand\smallO{
  \mathchoice
    {{\scriptstyle\mathcal{O}}}
    {{\scriptstyle\mathcal{O}}}
    {{\scriptscriptstyle\mathcal{O}}}
    {\scalebox{.7}{$\scriptscriptstyle\mathcal{O}$}}
}
 
\begin{table*}[!ht]
\begin{center}
\caption{
    \textbf{Computational complexity and memory requirements of different methods.}
    We report multiply-accumulate operations (MACS) as well as memory requirements
    (per layer) for batch size $b$, image size $s \times s \times c$,
    kernel size $k \times k$ and number of inner iterations $t$.
    We use $C= b s^2 c^2 k^2$, $M = b s^2 c$ and $P = c^2 k^2$.
    For a detailed explanation on what is reported see \Cref{sec:theoretical}.
    For some explanation on how the entries of this table were derived, 
    see \Cref{sec:computational_complexity}.
}
\label{table:theoretical}
\begin{tabular}{l|lll|lll}

\hline \noalign{\smallskip}

Method  & \multicolumn{3}{c|}{Input Transformations} & \multicolumn{3}{c}{Parameter Transformations} \\
        & Operations & MACS $\OO(\cdot)$ & Memory     & Operations & MACS $\OO(\cdot)$ & Memory $\OO(\cdot)$ \\

\noalign{\smallskip} \hline \noalign{\smallskip}


\stdconv 
& \CONV     & $C$   & $M$ 
& -         & -     & $P$ \\

\AOL
& \CONV     & $C$   & $M$ 
& \CONV     & $c^3 k^4$ & $5P$ \\

\BCOP
& \CONV     & $C$   & $M$ 
& BnB \& \MATMUL{}s & $c^3 k t + c^3 k^3$ & $c^2 k t + c^2 k^3$ \\

\Cayley
& \FFT{}s \& \MV{}s   & $b s^2 c^2$   & $\frac{5}{2} M$
& \FFT{}s \& \INV{}s  & $s^2 c^3$                & $\frac{3}{2} s^2 c^2$ \\

\CPL
& \CONV{}s \& \ACT    & $2C$ & $3M$
& \powermeth & $s^2 c^2 k^2$  & $P + s^2 c$ \\

\LOT
& \FFT{}s \& \MV{}s        & $b s^2c^2$ & $3M$
& \FFT{}s \& \MATMUL{}s    & $4s^2 c^3 t$ & $4 s^2 c^2 t$ \\

\SLL
& \CONV{}s \& \ACT  & $2C$              & $3M$
& \CONV{}s         & $c^3 k^4$    & $5P$ \\

\SOC 
& \CONV{}s    & $Ct_1$ & $Mt_1$
& F4                & $c^2 k^2 t_2$ &  $P$ \\ 


            
\noalign{\smallskip} \hline \noalign{\smallskip}
\end{tabular}
\end{center}
\end{table*}

%% file: sec/4_comparison.tex
\section{Theoretical Comparison} 
\label{sec:theoretical}
As illustrated in the last section, various ideas and methods have been proposed to parameterize \ols{} layers. This causes the different methods to have very different properties and requirements.
This section aims at highlighting the properties of the different algorithms, focusing on the algorithmic complexity and the required memory.
\medskip

\Cref{table:theoretical} provides an overview of the computational complexity and memory requirements
for the different layers considered in the previous section.
For the sake of clarity, the analysis is performed by considering separately the transformations applied to the input of the layers and those applied to the weights to ensure the \ols constraint.
Each of the two sides of the table contains three columns: 
\begin{enumerate*}[label=\roman*)]
\item \textit{Operations} contains the most costly transformations applied to the input as well
as to the parameters of different layers;
\item \textit{MACS} reports the computational complexity expressed in multiply-accumulate operations (MACS) involved in the transformations (only leading terms are presented);
\item \textit{Memory} reports the memory required by the transformation during the training phase.
\end{enumerate*}
\medskip

At training time, both input and weight transformations are required, thus the training complexity of the forward pass can be computed as the sum of the two corresponding MACS columns of the table. Similarly, the training memory requirements
can be computed as the sum of the two corresponding Memory columns of the table.
For the considered operations, 
the cost of the backward pass during training has the same computational complexity as the forward pass, and therefore increases the overall complexity by a constant factor.
At \inference time, all the parameter transformations can be computed just once and cached afterward. Therefore, the \inference complexity is equal to the complexity due to the input transformation (column 3 in the table).
The memory requirements at \inference time are much lower than those needed at the training time since intermediate activation values do not need to be stored in memory,  hence we do not report them in \Cref{table:theoretical}.
\medskip

Note that all the terms reported in \Cref{table:theoretical} depend on the batch size $b$,
the input size $s \times s \times c$,
the number of inner iterations of a method $t$,
and the kernel size $k \times k$.
(Often, $t$ is different at training and \inference time.)
%
For the sake of clarity, the MACS of a naive convolution implementation is denoted by $C$ ($C= b s^2 c^2 k^2$), the number of inputs of a layer is denoted by $M$ ($M = b s^2 c$), and the size of the kernel of a standard convolution is denoted by $P$ ($P = c^2 k^2$). 
Only the leading terms of the computations are reported in \Cref{table:theoretical}.
In order to simplify some terms, we assume that $c > \log_2 (s)$ and that rescaling a tensor 
(by a scalar) as well as
adding two tensors does not require any memory 
in order to do backpropagation.
We also assume that each additional activation does require extra memory.
All these assumptions have been verified to hold within 
\textit{PyTorch}, \cite{pytorch}. 
Also, when the algorithm described in the paper and the version provided in the supplied code differed, we considered the algorithm implemented in the code.
\medskip

The transformations reported in the table are
convolutions (\CONV),
Fast Fourier Transformations (\FFT),
matrix-vector multiplications (\MV),
matrix-matrix multiplications (\MATMUL),
matrix inversions (\INV),
as well as applications of an activation function (\ACT).
The application of algorithms such as
\BnB (BnB), \powermeth, and Fantastic 4 (F4) is also reported 
(see \Cref{sec:bounding_methods} for descriptions). 

\subsection{Analysis of the computational complexity}
It is worth noting that the complexity of the input transformations 
(in \Cref{table:theoretical}) is similar for all methods.
This implies that a similar scaling behaviour is expected at \inference time for the models.
\Cayley and \LOT apply an FFT-based convolution
and have computational complexity independent of the kernel size.
\CPL and \SLL require two convolutions, which make
them slightly more expensive at inference time.
Notably, \SOC requires multiple convolutions, 
making this method more expensive at \inference time.
\medskip

At training time, parameter transformations need to be applied in addition to the input transformations during every forward pass.
For \SOC and \CPL, the input transformations always dominate the parameter transformations in terms of computational complexity.
This means the complexity scales like $c^2$,
just like a regular convolution, with a further factor of $2$ and $5$ respectively.
All other methods require parameter transformations
that scale like $c^3$, making them more expensive for larger architectures.
In particular, we do expect \Cayley and \LOT to require long training times for larger models, since the complexity of their parameter transformations
further depends on the input size. 
\subsection{Analysis of the training memory requirements}

The memory requirements of the different layers are important, since they determine the maximum 
batch size
and the type of models we can train on a particular infrastructure.
At training time, typically all intermediate results are kept
in memory to perform backpropagation. This includes intermediate results for both input and parameter transformations.
The input transformation usually preserves the size,
and therefore the memory required is usually of $\mathcal{O}(M)$.
%
Therefore, for the input transformations, all methods require memory
not more than a constant factor worse than standard convolutions,
with the worst method being \SOC, with a constant $t_1$, typically equal to $5$. 
\medskip

In addition to the input transformation, we also need to store intermediate results of the parameter transformations in memory in order to evaluate the gradients.
Again, most methods approximately preserve the sizes during the parameter transformations, and therefore the memory required is usually of order $\mathcal{O}(P)$.
Exceptions to this rule are \Cayley and \LOT, which contain a much larger $\mathcal{O}(s^2 c^2)$ term, as well as \BCOP.

%% file: sec/5a_experiments.tex
\section{Experimental Setup}
\label{sec:methodology}
This section presents an experimental study aimed at comparing the performance of the considered layers with respect to different metrics. %
Before presenting the results, we first summarize the setup used in our experiments. For a detailed description see \cref{sec:experiments_appendix}.
To have a fair and meaningful comparison among the various models, all the proposed layers have been evaluated using the same architecture, loss function, and optimizer.
Since, according to the data reported in \Cref{table:theoretical}, different layers may have different throughput, to have a fair comparison with respect to the tested metrics,  we limited the total training time instead of fixing the number of training epochs. Results are reported for training times of 2h, 10h, and 24h on one A100 GPU.
\medskip

Our architecture is a standard convolutional network that doubles the number of channels whenever
the resolution is reduced \cite{cayley, brau}.
For each method, we tested architectures of different sizes. 
We denoted them as XS, S, M and L, depending on the number of parameters,
according to the criteria in \Cref{table:sizes}, ranging from 1.5M to 100M parameters.
\medskip

Since different methods benefit from different learning rates and weight decay, for each setting 
(model size, method and dataset), 
we used the best values resulting from a random search performed on multiple training runs on a validation set composed of $10\%$ of the original training set. More specifically, $16$ runs were performed for each configuration of randomly sampled hyperparameters, and we selected the configuration maximizing the \cra \wrt $\epsilon = 36/255$ (see \Cref{sec:greedy-search} for details).
\medskip

The evaluation was carried out using three different datasets: \CIFARS, \CIFARL \cite{cifar}, and \TinyIN \cite{tiny-imagenet}. 
Augmentation was used during the training (Random crops and flips on \CIFARS and \CIFARL, and \emph{RandAugment} \cite{cubuk2020randaugment} on \TinyIN). We use the loss function proposed by \cite{aol},  with the margin set to $2 \sqrt{2} \epsilon$, and temperature $0.25$.

\subsection{Metrics} \label{sec:metrics}
All the considered models were evaluated based on three main metrics: the \emph{throughput}, the required memory, and the \cra.

\paragraph{Throughput and epoch time}
The \emph{throughput} of a model is the average number of examples that the model can process per second. It determines how many epochs are processed in a given time frame. 
The evaluation of the throughput was performed on an 80GB-A100-GPU based on the average time of 100 mini-batches.
We measured the inference throughput with cached parameter transformations.

\paragraph{Memory required}
Layers that require less memory allow for larger batch
size, and the memory requirements also determine the 
type of hardware we can train a model on.
For each model, we measured and reported the maximal GPU memory occupied by tensors using the function \texttt{torch.cuda.max\_memory\_allocated()}
provided by the \emph{PyTorch} framework.
This is not exactly equal to the overall GPU memory requirement but gives a fairly good approximation of it.
Note that the model memory measured in this way also includes
additional memory required by the optimizer (e.g. to store the momentum term) as well as by the activation layers in the forward pass. However, this additional memory should be at most of order $\mathcal{O}(M+P)$.
As for the throughput, we evaluated and cached all calculations independent of the input at \inference time.

\paragraph{\Cra}
In order to evaluate the performance of a \ols network,
the standard metric is the \emph{\cra}.
An input is classified certifiably robustly with radius $\epsilon$ by a model, if no perturbations of the input with norm bounded by $\epsilon$ can change the prediction of the model.
\Cra measures the proportion of examples that are classified 
correctly as well as certifiably robustly.
For \ols models, a lower bound of the certified $\epsilon$-robust accuracy is the ratio of correctly classified inputs such that $\MM_f(x_i,l_i)>\epsilon\sqrt{2}$ 
where the \emph{margin} $\MM_f(x,l)$ 
of a model $f$ at input $x$ with label $l$,
given as $\MM_f(x,l) = f(x)_l-\max_{j\ne l} f_j(x)$,
is the difference
between target class score and the highest score of a different class.
For details, see \cite{lipschitzmargintraining}.

%
%


%% file: sec/5b_results.tex
\section{Experimental Results}  
\label{sec:results}

This section presents the results of the comparison performed by applying the methodology discussed in \Cref{sec:methodology}. The results related to the different metrics are discussed in dedicated subsections and the key takeaways are summarized in the radar-plot illustrated in \Cref{fig:radar_plot}.

\subsection{Training and inference times}
\input{fig/train_times}

\input{fig/val_times}

\Cref{fig:trainingtimes} plots the training time per epoch of the different models as a function of their size, while \Cref{fig:validationtimes} plots the corresponding inference throughput for the various sizes as described in \Cref{sec:methodology}.
As described in \Cref{table:architecture},
the model base width, referred to as $w$,
is doubled from one model size to the next.
We expect the training and inference time to scale with $w$ similarly to how individual layers scale with their number of channels, $c$ (in \Cref{table:theoretical}). This is because the width of each of the 5 blocks of our architecture is a constant multiple of the base width, $w$.

\paragraph{The training time} 
increases (at most) about linearly with $w$ for standard convolutions, whereas the computational complexity of each single convolution scales like $c^2$.
This suggests that parallelism on the GPU and the overhead from other operations 
(activations, parameter updates, etc.) 
are important factors determining the training time.
This also explains why \CPL (doing two convolutions, with identical kernel parameters) is only slightly
slower than a standard convolution, and \SOC (doing 5 convolutions) is only about 3 times slower than the standard convolution.
The \AOL and \SLL methods also require times comparable to a standard convolution for small models, although eventually, the $c^3$ term in the computation of the rescaling makes them slower for larger models.
Finally, \Cayley, \LOT, and \BCOP methods
take much longer training times per epoch.
For \Cayley and \LOT this behavior was expected, as they have a large $\Ord{s^2 c^3}$ term in their computational complexity. See \Cref{table:theoretical} for further details.

\paragraph{At \inference time} transformations of the weights are cached, therefore some methods (\AOL, \BCOP) do not have any overhead
compared to a standard convolution.
As expected, other methods (\CPL, \SLL, and \SOC) that apply additional convolutions to the input suffer from a corresponding overhead.
Finally, \Cayley and \LOT have a slightly different throughput due to their FFT-based convolution.
Among them, \Cayley is about twice as fast because it involves a real-valued FFT rather than a complex-valued one.
From \Cref{fig:validationtimes}, it can be noted that cached \Cayley and \CPL have the same inference time, 
even though CPL uses twice the number of convolutions. We believe this is due to the fact that the conventional FFT-based convolution is quite efficient for large filters, but PyTorch implements a faster algorithm, \ie, \textit{Winograd}, \cite{lavin2016fast}, that can be up to $2.5$ times faster.

\subsection{Training memory requirements}
\input{fig/train_val_memory}

The training and \inference memory requirements of the various models (measured as described in \Cref{sec:metrics})
are reported in \Cref{fig:memory_required} as a function of the model size.
The results of the theoretical analysis reported in \Cref{table:theoretical} suggest that the training memory requirements always have a term
linear in the number of channels, $c$ 
(usually the activations from the forward pass),
as well as a term quadratic in $c$ 
(usually the weights and all transformations applied to the weights
during the forward pass).
This behavior can also be observed from \Cref{fig:memory_required}.
For some of the models, the memory required approximately doubles from one model size to the next one, just like the width.
This means that the linear term dominates (for those sizes),
which makes those models relatively cheap to scale up.
%
For the \BCOP, \LOT, and \Cayley methods, the larger coefficients in the $c^2$ term 
(for \LOT and \Cayley the coefficient is even dependent on the input size, $s^2$)
cause this term to dominate.
This makes it much harder to scale those methods to more parameters. 
Method \LOT requires huge amounts of memory, in particular \LOT-L is too large to fit in 80GB GPU memory.
\medskip

Note that at test time, the memory requirements are much lower, because the intermediate activation values do not need to be stored, as there is no backward pass.
Therefore, at \inference time, most methods require a very
similar amount of memory as a standard convolution.
The \Cayley and \LOT methods require more memory since perform the calculation in the Fourier space, as they create an intermediate representation of the weight matrices of size $\Ord{s^2 c^2}$.
\subsection{\Cra}
\begin{figure*}
\centering
\includegraphics[scale=0.5]{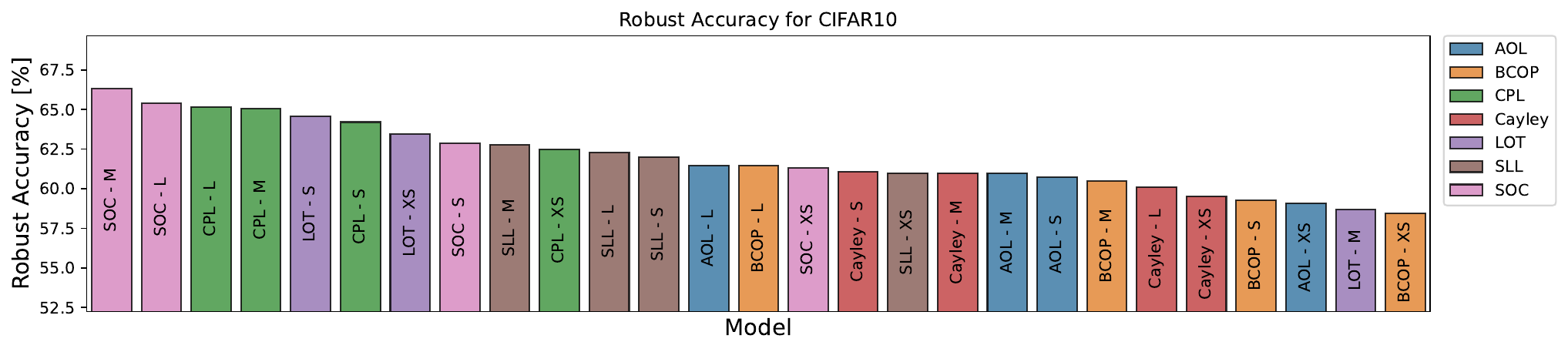}
\includegraphics[scale=0.5, clip, trim=0 0 2.8cm 0]{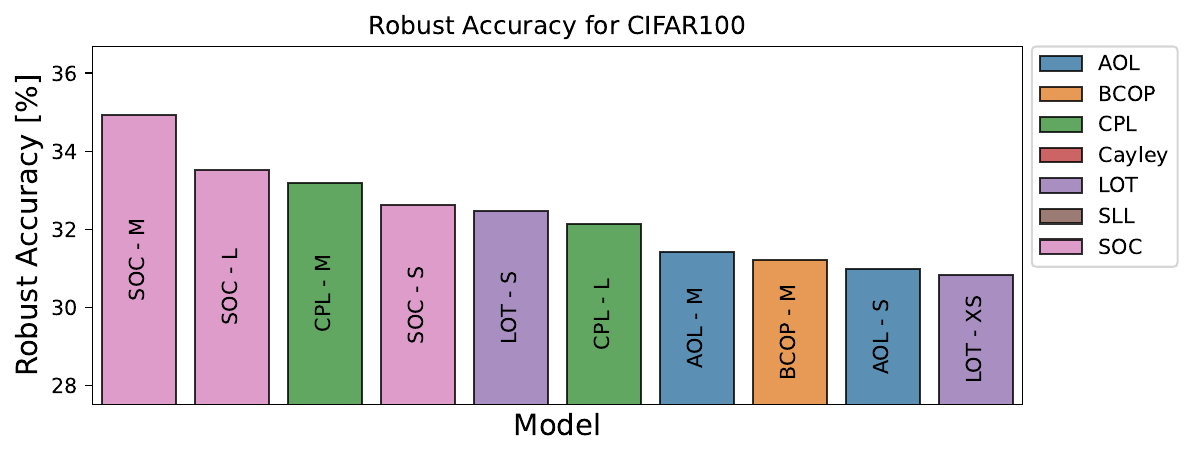}%
\includegraphics[scale=0.5, clip, trim=0 0 2.8cm 0]{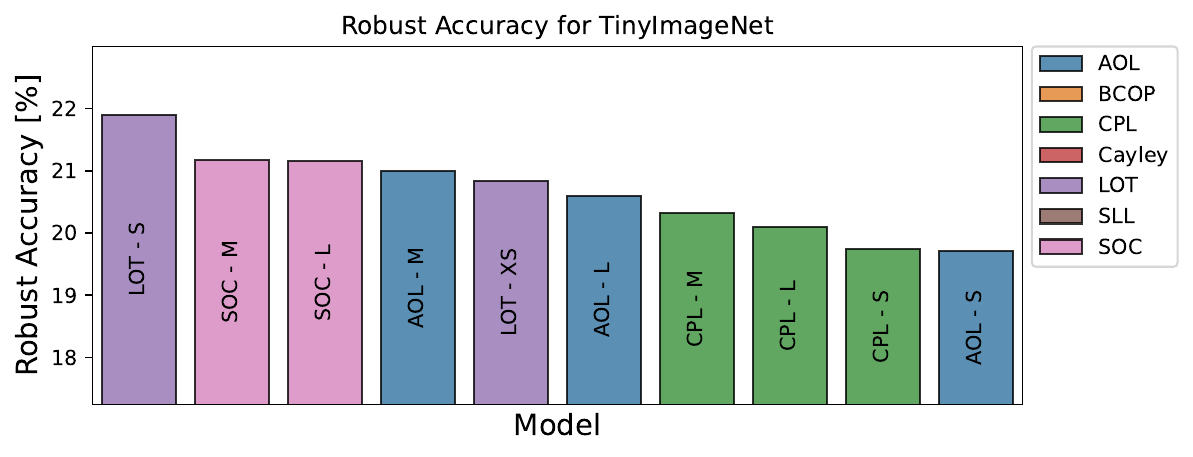}
\caption{
    \textbf{\Cra} by decreasing order.
    Note that the axes do not start at 0.
    For \CIFARL and \TinyIN only the 10 best performing
    models are shown.}
\label{fig:barplot}
\end{figure*}
\input{tab/cr_table_24h}

The results related to the accuracy and the \cra for the different methods, model sizes,
and datasets measured on a $24$h training budget are summarized in \Cref{tab:cr24h}. The differences among the various model sizes are also highlighted in \Cref{fig:barplot} by reporting the sorted values of the \cra.
Further tables and plots relative to different training budgets can be found in \Cref{sec:further_results}.
The reader can compare our results with the state-of-the-art \cra summarized in \Cref{sec:sota}. However, it is worth noting that, to reach state-of-the-art performance, authors often carry out experiments using large model sizes and long training times, which makes it hard to compare the methods themselves. On the other hand, the evaluation proposed in this paper allows a fairer comparison among the different methods, since it also considers timing and memory aspects.
%
This restriction based on time, rather than the number of epochs, ensures that merely enlarging the model size does not lead to improved performance, as bigger models typically process fewer epochs of data.
Indeed, in our results in \Cref{fig:barplot} it is usually
the M (and not the L) model that performs best.
To assign a score that combines the performance 
of the methods 
over all the three datasets,
we sum the number of times that each method is ranked in the first position, in the top-3, and top-10 positions. In this way, top-1 methods are counted three times, and top-3 methods are counted twice.
The scores in the radar-plot shown in \Cref{fig:radar_plot} are based on those values.
\medskip

Among all methods, \SOC achieved a top-1 robust accuracy twice and a top-3 one 6 times, outperforming all the other methods.
\CPL ranks twice in the top-3 and 9 times in the top-10 positions, showing that it generally has a more stable performance compared with other methods. 
\LOT achieved the best \cra on \TinyIN, appearing further $5$ times in the top-10.
\AOL did not perform very well on \CIFARS,
but reached more competitive results on \TinyIN,
ending up in the top-10 a total of $5$.
An opposite effect can be observed for \SLL,
which performed reasonably well on \CIFARS, 
but not so well on the two datasets with more classes, placing in the top-10 only once.
This result is tied with \BCOP, which also
has only one model in the top-10.
Finally, \Cayley is consistently outperformed by the other methods.
The very same analysis can be applied to the clean accuracy, whose sorted bar-plots are reported in \Cref{sec:further_results}, where the main difference is that \Cayley performs slightly better for that metric.
Furthermore, it is worth highlighting that \CPL is sensitive to weight initialization. We faced numerical errors during the $10$h and $24$h training of the small model on \CIFARL.

%% file: fig/train_times.tex
\begin{figure}[b]
    \centering
    \includegraphics[width=\columnwidth]{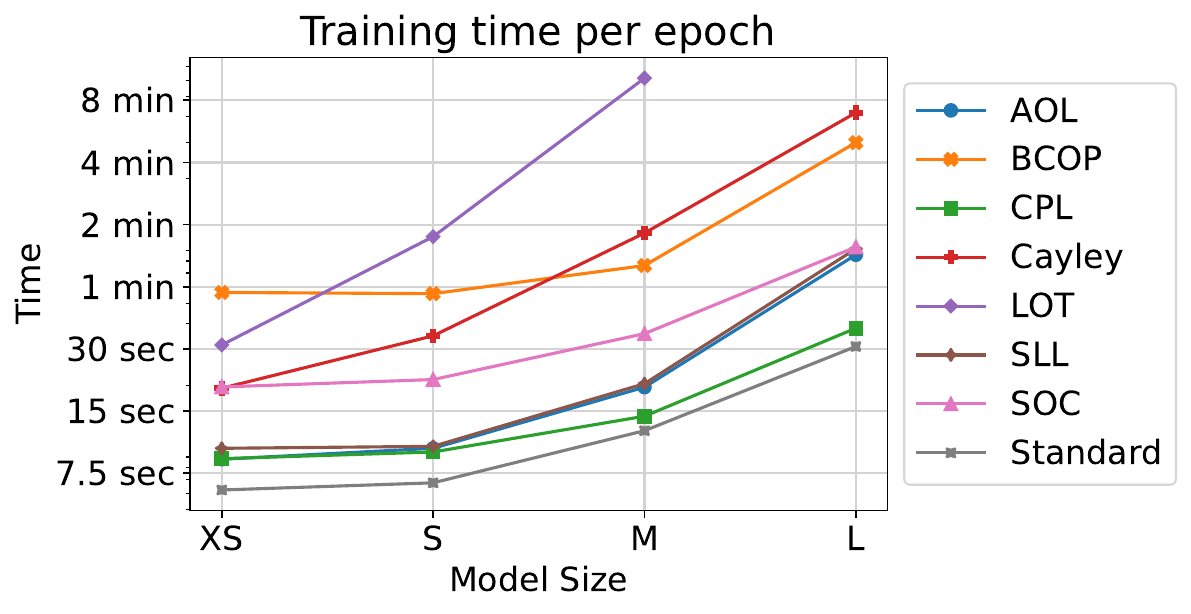}
    
    \caption{
        \textbf{Training time} per epoch (on \CIFARS) 
        for different methods and different model sizes.
    }
    \label{fig:trainingtimes}
\end{figure}

%% file: fig/val_times.tex
\begin{figure}[b]
    \centering
    \includegraphics[width=\columnwidth]{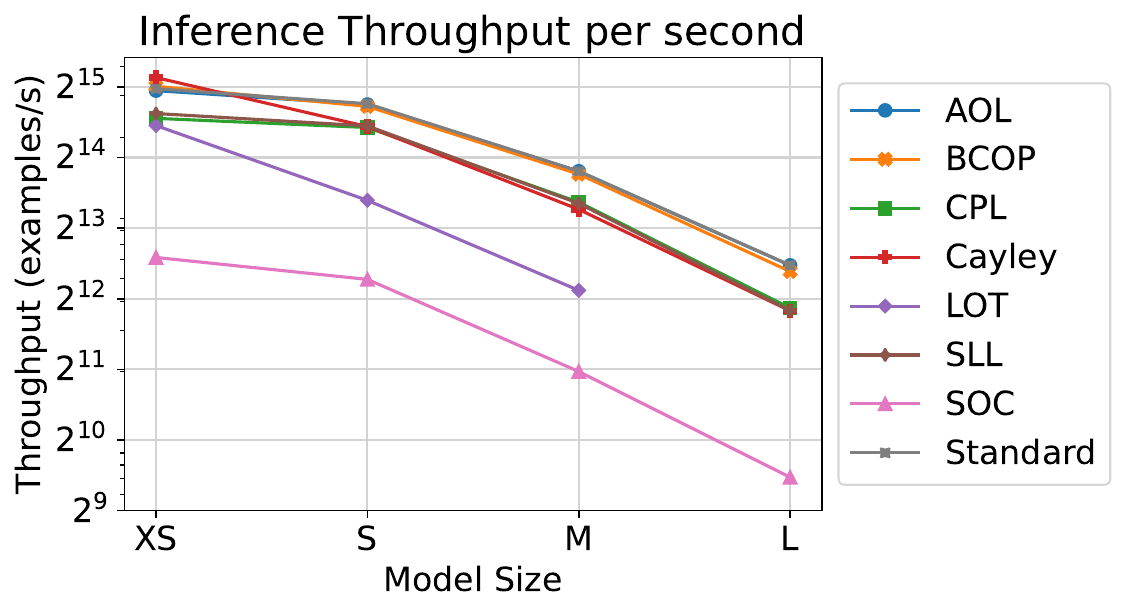}    
    
    \caption{
        \textbf{\Inference throughput}
        for different methods as a function of their size for \CIFARS sizes input images.
        All parameter transformations
        have been evaluated and cached beforehand
        }
    \label{fig:validationtimes}
\end{figure}

%% file: fig/train_val_memory.tex
\begin{figure}
    \centering

    \includegraphics[width=\columnwidth]{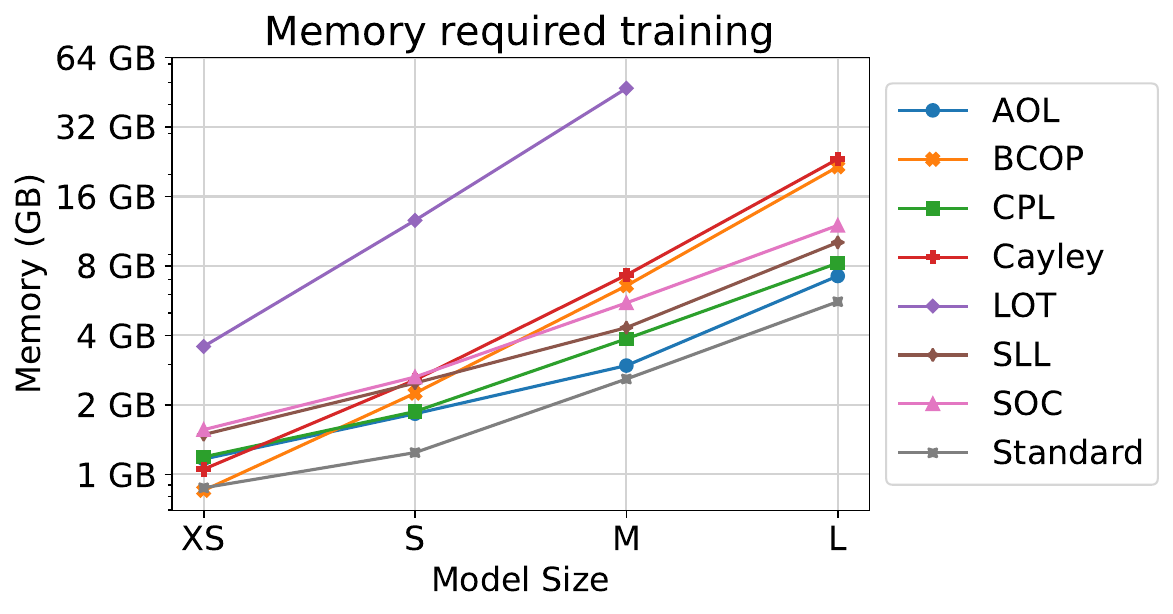}    
    \includegraphics[width=\columnwidth]{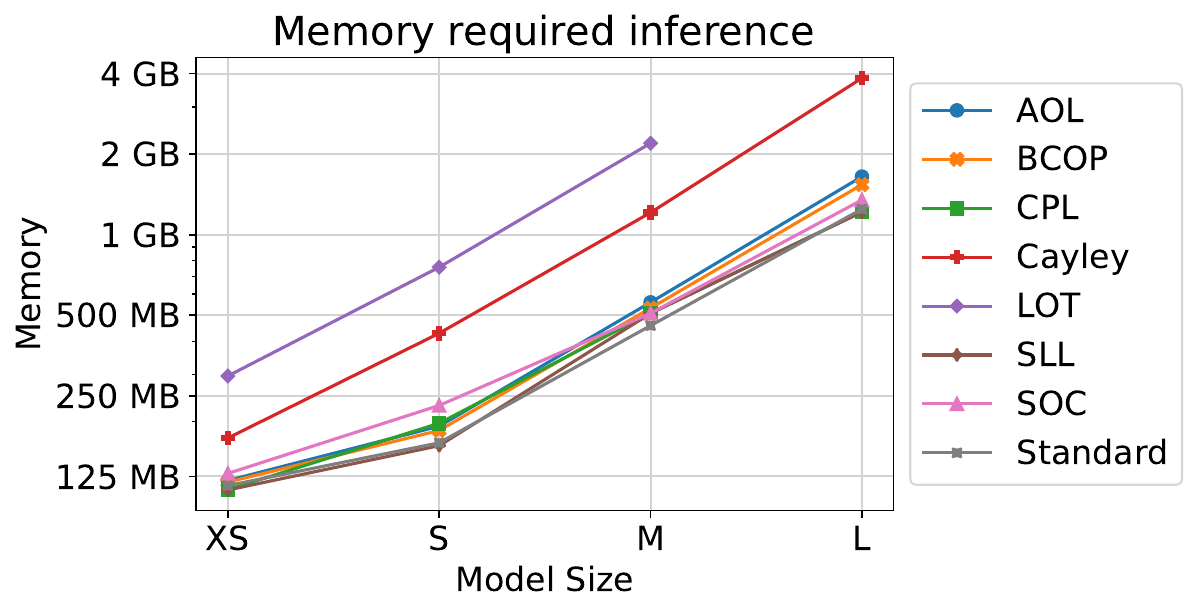}
    
    \caption{
        \textbf{Memory required} at training and \inference time for input size $32\times 32$.
    }
    \label{fig:memory_required}
\end{figure}

%% file: tab/cr_table_24h.tex
\begin{table}[t]
\setlength{\tabcolsep}{4pt}
\centering
\caption{\Cra for radius $\epsilon=36/255$ on the evaluated datasets. Training is performed for $24$ hours.}
\label{tab:cr24h}

\resizebox{\columnwidth}{!}{\input{tab/raws/CR_table_24h}}
\end{table}



%% file: tab/raws/CR_table_24h.tex
\begin{tabular}{lcccc|cccc}
\toprule
 & \multicolumn{4}{c|}{Accuracy [\%]} & \multicolumn{4}{c}{Robust Accuracy [\%]} \\
Methods & XS & S & M & L & XS & S & M & L \\
\midrule
\midrule
\multicolumn{9}{c}{\textbf{\CIFARS}} \\
\midrule
\textbf{AOL} & 71.7 & 73.6 & 73.4 & 73.7 & 59.1 & 60.8 & 61.0 & \textbf{61.5} \\
\textbf{BCOP} & 71.7 & 73.1 & 74.0 & 74.6 & 58.5 & 59.3 & 60.5 & \textbf{61.5} \\
\textbf{CPL} & 74.9 & 76.1 & 76.6 & 76.8 & 62.5 & 64.2 & 65.1 & \textbf{65.2} \\
\textbf{Cayley} & 73.1 & 74.2 & 74.4 & 73.6 & 59.5 & \textbf{61.1} & 61.0 & 60.1 \\
\textbf{LOT} & 75.5 & 76.6 & 72.0 & - & 63.4 & \textbf{64.6} & 58.7 & - \\
\textbf{SLL} & 73.7 & 74.2 & 75.3 & 74.3 & 61.0 & 62.0 & \textbf{62.8} & 62.3 \\
\textbf{SOC} & 74.1 & 75.0 & 76.9 & 76.9 & 61.3 & 62.9 & \textbf{66.3} & 65.4 \\
\midrule
\midrule
\multicolumn{9}{c}{\textbf{\CIFARL}} \\
\midrule
\textbf{AOL} & 40.3 & 43.4 & 44.3 & 41.9 & 27.9 & 31.0 & \textbf{31.4} & 29.7 \\
\textbf{BCOP} & 41.4 & 42.8 & 43.7 & 42.2 & 28.4 & 30.1 & \textbf{31.2} & 29.2 \\
\textbf{CPL} & 42.3 & - & 45.2 & 44.3 & 30.1 & - & \textbf{33.2} & 32.1 \\
\textbf{Cayley} & 42.3 & 43.9 & 43.5 & 42.9 & 29.2 & \textbf{30.5} & 30.5 & 29.5 \\
\textbf{LOT} & 43.5 & 45.2 & 42.8 & - & 30.8 & \textbf{32.5} & 29.6 & - \\
\textbf{SLL} & 41.4 & 42.8 & 42.4 & 42.1 & 28.9 & \textbf{30.5} & 29.9 & 29.6 \\
\textbf{SOC} & 43.1 & 45.2 & 47.3 & 46.2 & 30.6 & 32.6 & \textbf{34.9} & 33.5 \\
\midrule
\midrule
\multicolumn{9}{c}{\textbf{\TinyIN}} \\
\midrule
\textbf{AOL} & 26.6 & 29.3 & 30.3 & 30.0 & 18.1 & 19.7 & \textbf{21.0} & 20.6 \\
\textbf{BCOP} & 22.4 & 26.2 & 27.6 & 27.0 & 13.8 & 16.9 & \textbf{17.2} & 16.8 \\
\textbf{CPL} & 28.3 & 29.3 & 29.8 & 30.3 & 18.9 & 19.7 & \textbf{20.3} & 20.1 \\
\textbf{Cayley} & 27.8 & 29.6 & 30.1 & 27.2 & 17.9 & \textbf{19.5} & 19.3 & 16.7 \\
\textbf{LOT} & 30.7 & 32.5 & 28.8 & - & 20.8 & \textbf{21.9} & 18.1 & - \\
\textbf{SLL} & 25.1 & 27.0 & 26.5 & 27.9 & 16.6 & 18.4 & 17.7 & \textbf{18.8} \\
\textbf{SOC} & 28.9 & 28.8 & 32.1 & 32.1 & 18.9 & 18.8 & \textbf{21.2} & 21.1 \\

\bottomrule
\end{tabular}

%% file: sec/7_Conclusion.tex
\section{Conclusions and Guidelines}
This work presented a comparative study of state-of-the-art 1-Lipschitz layers under the lens of different metrics, such as time and memory requirements, accuracy, and \cra, all evaluated at training and \inference time. A theoretical comparison of the methods in terms of time and memory complexity was also presented and validated by experiments.
\medskip

Taking all metrics into account (summarized in \Cref{fig:radar_plot}), the results are in favor of CPL, due to its highest performance and lower consumption of computational resources.
When large computational resources are available and the application does not impose stringent timing constraints during \inference and training, the \SOC layer could be used, due to its slightly better performance.
Finally, those applications in which the \inference time is crucial may take advantage of \AOL or \BCOP, which do not introduce additional runtime overhead (during \inference) 
compared to a standard convolution.

%% file: sec/9_appendix.tex
\begin{center}
  \Large
  \textbf{Technical Appendix of ``\Title''}
\end{center}
\vspace{1pt}
\vspace{10pt}

\input{appendix/1_algorithms_for_bounded_spectral_norm}

\input{appendix/2_algorithms_omitted}

\input{appendix/3_justification_for_theoretical_results}

\input{appendix/4_comparison_with_sota}
\input{appendix/5_experimental_setup}

\input{appendix/9_other}

\input{appendix/6_further_results}
\input{appendix/7_reported_issues}
\input{appendix/8_code}
\clearpage
\input{tab/cr_table_cifar10_full}
\input{tab/cr_table_cifar100_full}
\input{tab/cr_table_tinyimagenet_full}
\input{tab/cr_table_cifar10_full_1x1}

%% file: appendix/1_algorithms_for_bounded_spectral_norm.tex
\section{Spectral norm and orthogonalization}
\label{sec:bounding_methods}
A lot of recently proposed methods do rely on a way
of parameterizing orthogonal matrices or 
parameterizing matrices with bounded spectral norm.
%
%
We present methods that are frequently used below:

\paragraph{Bjorck \& Bowie \cite{Bjorck_1971_SIAM}} \label{sec:bnb}
introduced an iterative algorithm that finds the closest orthogonal matrix to the given input matrix.
In the commonly used form, this is achieved 
by computing a sequence of matrices using
\begin{equation}
    A_{k+1} = A_k\left(I + \frac12 Q_k \right),
    \quad\mbox{for }
    Q_k=I-A_k^\top A_k
\end{equation}
where $A_0=A$, is the input matrix.
The algorithm is usually truncated after a fixed
number of steps, during training often $3$ iterations
are enough, and for \inference more (e.g. $15$) iterations are used to ensure a good approximation.
Since the algorithm is differentiable, it can 
be applied to construct \ols networks as proposed initially in \cite{anil_fc} or also as an auxiliary method for more complex strategies \cite{bcop}.
%
%



\paragraph{Power Method} \label{sec:power_method}
The \emph{power method} 
was used in \cite{miyato}, \cite{GloRo} and \cite{CPL} in order to bound the
spectral norm of matrices. 

It starts with a random initialized vector $\vec{u}_0$,
and iteratively applies the following:
\begin{equation} \label{eq:powermethod}
    \vec{v}_{k+1} = \frac{W^\top \vec{u}_k}{\|W^\top \vec{u}_k\|_2}, \quad
    \vec{u}_{k+1} = \frac{W \vec{v}_{k+1}}{\|W \vec{v}_{k+1}\|_2}.
\end{equation}
Then the sequence $\sigma_k$ converges to the spectral norm of $W$,
for $\sigma_k$ given by
\begin{equation}
    \sigma_k = \vec{u}_k^\top W \vec{v}_k.
\end{equation}
This procedure allows us to obtain the spectral norm of matrices,
but it can also be efficiently extended to find the spectral norm
of the Jacobian of convolutional layers.
This was done for example by \cite{Farnia_2018_ICLR,GloRo},
%
using the fact that 
the transpose of a convolution operation 
(required to calculate \Cref{eq:powermethod})
is a
convolution as well, with a kernel that can be constructed
from the original one by transposing the channel dimensions and
flipping the spatial dimensions of the kernel.

When the power method is used on a parameter matrix of a layer,
we can make it even more efficient with a simple trick.
We usually expect the parameter matrix to change only slightly
during each training step, so we can store the result $\vec{u}_k$ 
during each training step,
and start the power method with
this vector as $\vec{u}_0$
during the following training step.
With this trick it is enough to do a single iteration of the
power method at each training step.
The \powermeth is usually not differentiated through.

\paragraph{Fantasic Four} \label{sec:fantastic_four}
proposed, in \cite{fantastic4}, allows upper bounding the Lipschitz constant of a convolution.
The given bound is generally not tight, so using the method directly does not give good results.
Nevertheless, since various methods require a way of bounding the spectral norm to have convergence guarantees, Fantastic Four is often used.

%% file: appendix/2_algorithms_omitted.tex
\section{Algorithms omitted in the main paper}
\label{sec:omitted}
Observe that the strategies presented in 
\cite{parseval, orthDNN, miyato, lezcano,eco,ONI_2020_huang,sandwich_2023_wang} 
have intentionally not been compared for different reasons. In the works presented in \cite{parseval, orthDNN}, the Lipschitz constraint was solely used during training and no guarantees were provided that the resulting layers are \ols. The method proposed in \cite{miyato} has been extended by Fantastic 4 \cite{fantastic4}  and, indeed, can only be used as an auxiliary method to upper-bound the Lispchitz constant. The method proposed in \cite{lezcano} only works for linear layers and can be thought of as a special case of SOC 
(described in \Cref{sec:methods}). We will give detailed reasons for the other methods below.

\begin{figure}[h!]
    \centering
    \includegraphics[width=0.45\textwidth]{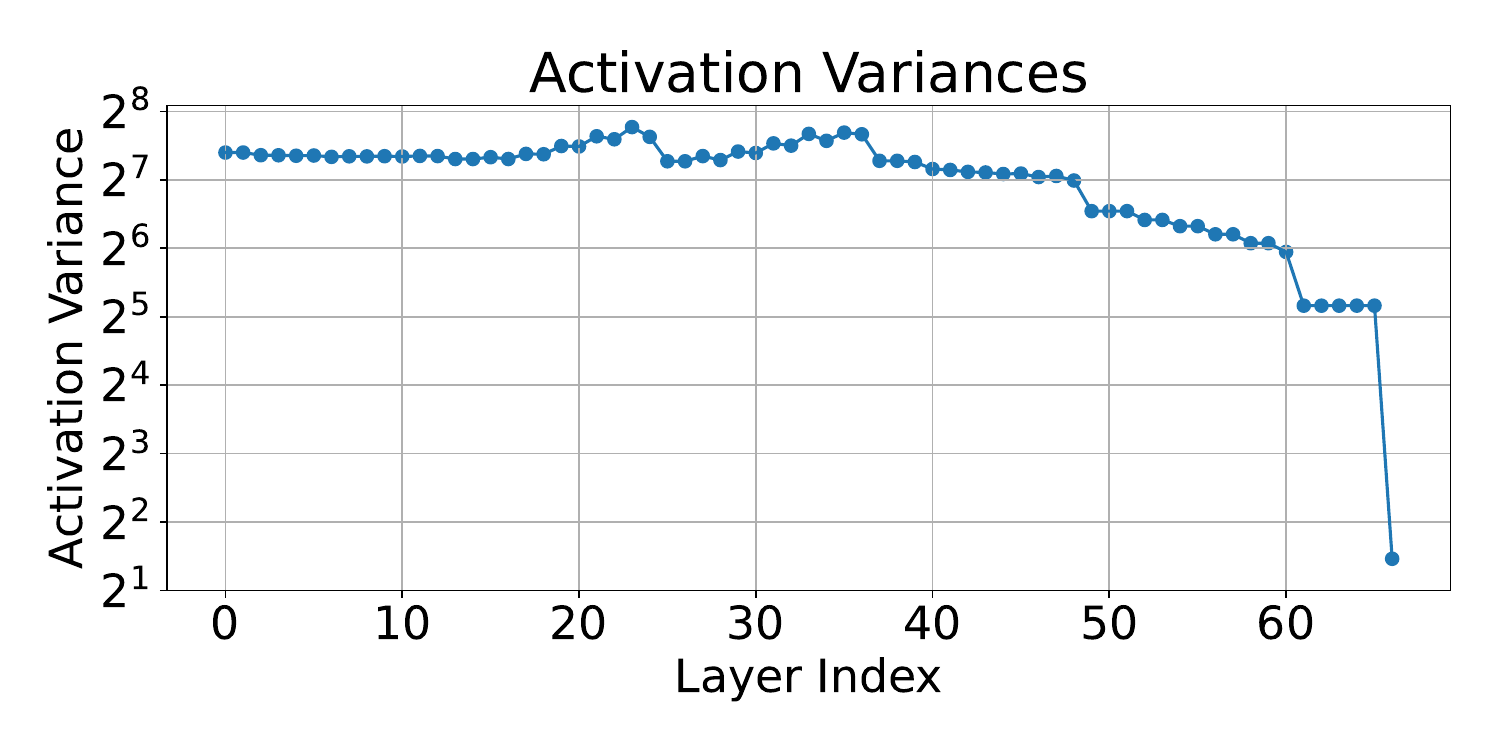}
    \includegraphics[width=0.45\textwidth]{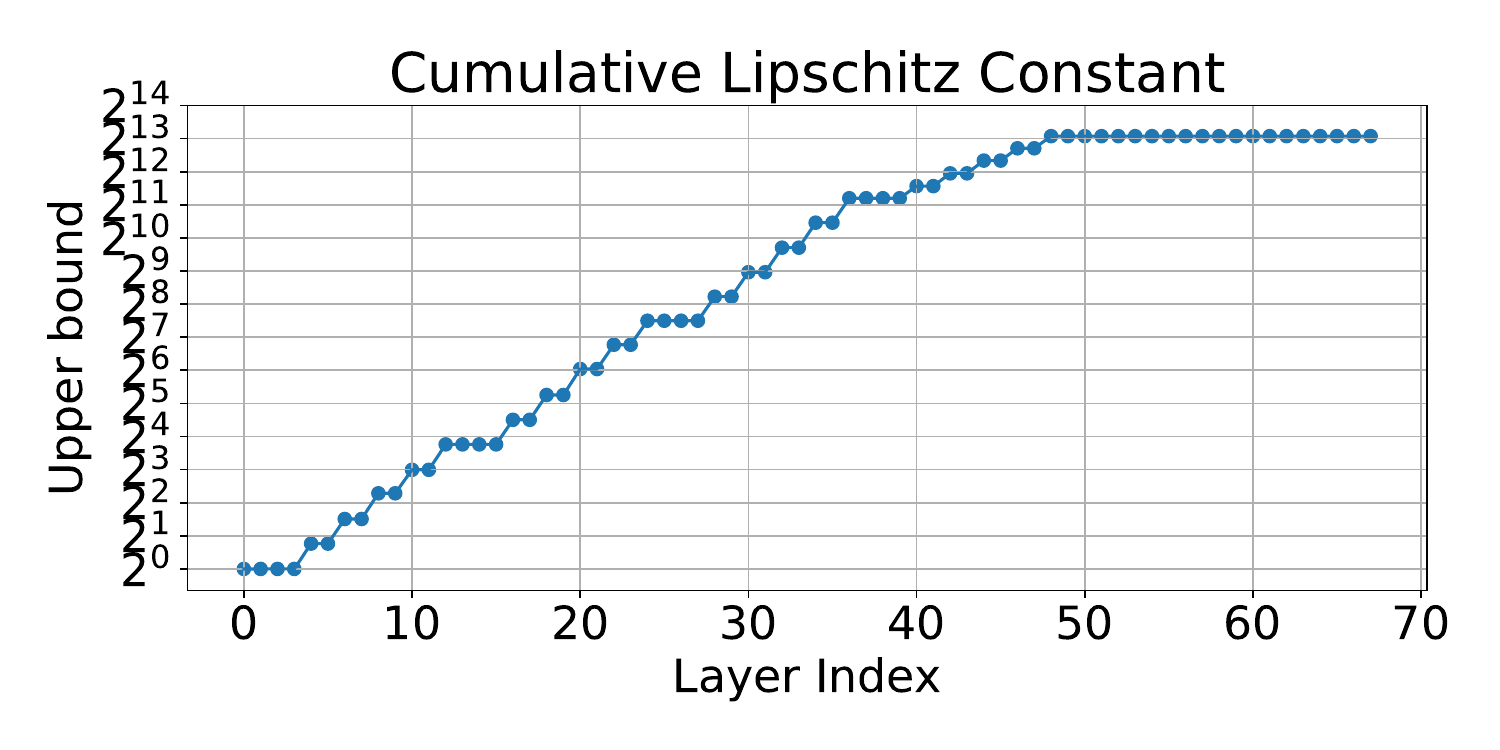}
    \caption{
        \textbf{Left:}
        Variance of a validation batch over the batch dimension.
        For \ols layers, this property should be non-increasing,
        as proven in \Cref{sec:BatchActVar}.
        \textbf{Right:}
        Upper bound on the Lipschitz Constant applying the power
        method to every linear layer, and multiplying the results.
        Plot for \ECO on \CIFARS, with model S.
    }
    \label{fig:actvareco}
\end{figure}

\paragraph{\ONI}
The method \ONI \cite{ONI_2020_huang} proposed the orthogonalization
used in \LOT.
They parameterize orthogonal matrices as $(V V^\top)^{-\frac{1}{2}}V$,
and calculate the inverse square root using Newton's iterations.
They use this methods to define \ols linear layers.
However, the extension to convolutions only uses a simple unrolling,
and does not provide a tight bound in general.
Therefore, we did not include the method in the paper.
\paragraph{ECO}
\emph{Explicitly constructed orthogonal (ECO)} convolutions \cite{eco} also do use properties of the Fourier domain in order to parameterize a convolution. However, they do not actually calculate the convolution in the Fourier domain, but instead parameterize a layer in the Fourier domain, and then use an inverse Fourier transformation to obtain a kernel from this parameterization.
We noticed, however, that the implementation provided
by the authors does not produce \ols layers
(at least with our architecture),
as can be seen in \Cref{fig:actvareco}.
There, we report the \emph{batch activation variance}
(defined in  \Cref{sec:BatchActVar})
as well as the spectral norm of each layer.
The \emph{batch activation variance} should be non-increasing for
\ols layers (also see \Cref{sec:BatchActVar}), however, for \ECO this is not the case.
Also, power iteration shows that the Lipschitz constant of individual
layers is not $1$.
Therefore we do not report this method in the main paper.



\paragraph{\Sandwich}
The authors of \cite{sandwich_2023_wang} introduced the \emph{Sandwich} layer.
It considers a layer of the form
\begin{align} \label{eq:sandwich}
    l(x) = \sqrt{2}A^T\Psi\ReLU{\sqrt{2}\Psi^{-1}Bx +b},
\end{align}
for $\sigma$ typically the ReLU activation.
The authors propose a (simultaneous) parameterization of $A$ and $B$, based on the Cayley Transform,
that guarantees the whole layer to be \ols.
They also extend the idea to convolutions.
However, for this they require to apply two Fourier
transformations as well as two inverse ones.
During the training of the models within the Sandwich layers, a severe vanishing gradient phenomena happens. We summarize the Frobenious norm of the gradient, obtained by inspecting the inner blocks during the training, in \Cref{table:zero_grad_sandwich}. For this reason we did not report the results in the main paper. 
\input{tab/zero_gradient}

%% file: tab/zero_gradient.tex
\setlength{\tabcolsep}{4pt}
\begin{table}[b]
\begin{center}
\caption{
    \textbf{Vanishing Gradient Phenomena of Sandwhich Layer}. ConvNetXS model has been tested with a small batch size (32). Training of deeper layers (i.e., layers that are close to the input of the network) is tough due to the almost zero gradients.}
\label{table:zero_grad_sandwich}
\begin{tabular}{rl|lr}
\noalign{\smallskip} \hline \noalign{\smallskip}
& Layer name & Output Shape & Gradient Norm\\
\noalign{\smallskip} \hline \noalign{\smallskip}



& First Conv ($1\times1$ kernel size)  & $(32,16,32,32)$ & \numprint{3.36e-7}    \\
& Activation                & $(32,16,32,32)$ & \numprint{3.36e-7}    \\

& Downsize Block($3$)            & $(32,32,16,16)$ & \numprint{3.79e-6} \\
& Downsize Block($3$)            & $(32,64,8,8)$ & \numprint{5.56e-5}    \\
& Downsize Block($3$)            & $(32,128,4,4)$ & \numprint{7.83e-4}    \\
& Downsize Block($3$)            & $(32,256,2,2)$ & \numprint{8.15e-3}    \\
& Downsize Block($1$)            & $(32,512,1,1)$ & \numprint{1.04e-1}    \\
& Flatten                   & $(32,512)$  & \numprint{1.04e-1}  \\
& Linear                    & $(32,512)$  & \numprint{1.82e-1}   \\
& First Channels             & $(32,10)$  & \numprint{1.82e-1}     \\

\noalign{\smallskip} \hline
\end{tabular}
\end{center}
\end{table}
\setlength{\tabcolsep}{1.4pt}

%% file: appendix/3_justification_for_theoretical_results.tex
\section{Computation Complexity and Memory Requirement}
\label{sec:computational_complexity}
In this section we give some intuition
of the values in \Cref{table:theoretical}.

Recall that we consider a layer with
input size $s \times s \times c$,
and kernel size $k \times k$,
batch size $b$,
and (for some layers) we will denote
the number of inner iterations by $t$.

We also use
$C=b s^2 c^2 k^2$ and $M=b s^2 c$ and $P = c^2 k^2$.


%

\paragraph{\AOL:}
In order to compute the rescaling matrix for \AOL,
we need to convolve the kernel with itself.
This operation has complexity $\mathcal{O}(c^3 k^4)$.
It outputs a tensor of size $c \times c \times (2k-1) \times (2k-1)$,
so in total we require memory of about 
$5P$ for the parameter as well as the transformation.

\paragraph{\BCOP:}
For \BCOP we only require a single convolution 
as long as we know the kernel.
However, we do require a lot of computation to
create the kernel for this convolution.
In particular, we require $2k-1$ matrix orthogonalizations
(usually done with \BnB),
as well as $\mathcal{O}(k^3)$ matrix multiplications for
building up the kernel.
These require about $c^3kt + c^3k^3$ MACS as well as $c^2kt + c^2 k^3$ memory.
\paragraph{\Cayley:}
Cayley Convolutions make use of the 
fact that circular padded convolutions are vector-matrix
products in the Fourier domain.
Applying the \fft to inputs and weights has complexity
of $\mathcal{O}(b c s^2 \log(s^2))$
and $\mathcal{O}(c^2 s^2 \log(s^2))$.
Then, we need to orthogonalize $\frac{1}{2} s^2$ matrices.
Note that the factor of $\frac{1}{2}$ appears due to
the fact that the Fourier transform of a real matrix
has certain symmetry properties, and we can use that
fact to skip half the computations.
Doing the matrix orthogonalization with the Cayley Transform 
requires taking the inverse of a matrix, 
as well as matrix multiplication,
the whole process has a complexity of 
about $s^2 c^3$.
%
The final steps consists of doing $\frac{1}{2}b s^2$
matrix-vector products, requiring 
$\frac{1}{2} b s^2 c^2$ \MACS,
as well as another \fft.
Note that under our assumption that $c > \log(s^2)$,
the \fft operation is dominated by other operations.
%
Cayley Convolutions require padding the kernel from a size of 
$c \times c \times k \times k$ to a (usually much larger) 
size of $c \times c \times s \times s$
requiring a lot of extra memory.
In particular we need to keep the output of the
(real) fast Fourier transform, the matrix inversion
as well as the matrix multiplication in memory,
requiring about $\frac{1}{2} s^2 c^2$ memory each.

\paragraph{\CPL:}
\CPL applies two convolutions as well as an activation
for each layer. 
They also use the Power Method (on the full convolution), 
however, its computational
cost is dominated by the application of the convolutions.

\paragraph{\LOT:}
Similar to \Cayley, \LOT performs the convolution
in Fourier space.
However, instead of using the Cayley transform,
they parameterize orthogonal matrices as 
$V(V^TV)^{-\frac12}$.
To find the inverse square root, authors relay on an iterative \textit{Newton Method}. In details, let $Y_0=V^TV$ and $Z_0=I$, then $Y_i$ defined as
\begin{equation}
    Y_{i+1} = \frac12 Y_i\left(3I-Z_iY_i\right),\quad Z_{i+1} = \frac12\left(3I-Z_iY_i\right)Z_i,
\end{equation}
converges to $(V^TV)^{-\frac12}$.
Executing this procedure,
includes computing $4 s^2 t$ matrix multiplications,
requiring about $4 s^2 c^3 t$ MACS as well as $4 s^2 c^2 t$ memory.

\paragraph{\SLL:}
Similar to \CPL, each \SLL layer also requires 
evaluating two convolutions as well as one activation.
However, \SLL also needs to compute the AOL rescaling,
resulting in total computational cost 
of $2C + \mathcal{O}(c^3 k^4)$.

\paragraph{\SOC:}
For each \SOC layer we require applying $t$ convolutions.
Other required operations (application of Fantastic 4 for an
initial bound, as well as parameterizing the kernel
such that the Jacobian is skew-symmetric)
are cheap in comparison.

%% file: appendix/4_comparison_with_sota.tex
\section{Comparison with SOTA} \label{sec:sota}
In this section, we report state-of-the-art results from the literature.
In contrast to our comparison, the runs reported often use larger
architectures and longer training times.
Find results in Table \cref{table:cifar10sota}.
\input{tab/cifar10_sota}
    

%% file: tab/cifar10_sota.tex
\begin{table}[t]
\begin{center}
\caption{
    \textbf{SOTA from the literature on \CIFARS} 
    sorted by publication date (from older to newer). Readers can note that there is a clear trend of increasing the model dimension to achieve higher robust accuracy.
}
\label{table:cifar10sota}
\begin{tabular}{lc|rrrr|r}
\toprule
 &  & \multicolumn{4}{c|}{Certifiable Accuracy [\%]} & Number of \\
Method & Std.Acc [\%] & $\varepsilon=\frac{36}{255}$ & $\varepsilon=\frac{72}{255}$ & $\varepsilon=\frac{108}{255}$ & $\varepsilon=1$ & Parameters \\
\midrule
\textbf{\BCOP Large} \cite{bcop}& 72.1 & 58.2 & - & - & - & 2M\\
\textbf{\Cayley KW-Large} \cite{cayley} & 75.3 & 59.1& - & - & - &2M\\
\textbf{\SOC LipNet-25} \cite{soc} & 76.4 & 61.9 & - & - & - & 24M\\
\textbf{\AOL Large}  \cite{aol}   & 71.6 & 64.0 & 56.4 & 49.0 & 23.7 & 136M \\
\textbf{\LOT LipNet-25} \cite{lot}& 76.8 & 64.4 & 49.8 & 37.3 &- & 27M\\
\textbf{\SOC LipNet-15 + CRC} \cite{singla2022improved} & 79.4 & 67.0 & 52.6 & 38.3 & - & 21M\\
\textbf{\CPL XL} \cite[Table1]{CPL}       & 78.5 & 64.4 & 48.0 & 33.0 & -    & 236M \\
\textbf{\SLL X-Large} \cite{SLL}  & 73.3 & 65.8 & 58.4 & 51.3 & 27.3 & 236M \\

\bottomrule
\end{tabular}
\end{center}
\end{table}

%% file: appendix/5_experimental_setup.tex
\section{Experimental Setup} \label{sec:experiments_appendix}
In addition to theoretically analyzing different
proposed layers, we also do an empirical
comparison of those layers.
In order to allow for a fair and meaningful
comparison, we try to fix the architecture,
loss function and optimizer, and evaluate
all proposed layers with the same setting.

From the data in \Cref{table:theoretical} we know that
different layers will have very different throughputs.
In order to have a fair comparison despite of that, 
we limit the total training time 
instead of fixing a certain amount of training epochs.
We report results of training for 2h, 10h as well as 24h.

We describe the chosen setting below.

\subsection{Architecture}

\input{tab/architecture}

\input{tab/conv_block}

We show the architecture used for our experiments
in \Cref{table:architecture,table:convblock}.
It is a standard convolutional architecture,
that doubles the number of channels whenever
the resolution is reduced.
Note that we exclusively use convolutions with
the same input and output size 
as an attempt to make the model less dependent 
on the initialization used by the convolutional layers.
We use kernel size $3$ in all our main experiments.
The layer 
\emph{Zero Channel Padding}
in \Cref{table:architecture}
just appends channels with value $0$ to the input, and the layer
\emph{First Channels($c$)} outputs only the first $c$ channels,
and ignores the rest.
Finally, the layer \emph{Pixel Unshuffle} (implemented in \textit{PyTorch})
takes each $2 \times 2 \times c$ patches of an image and reshapes them
into size $1 \times 1 \times 4c$.

For each \ols layer, we also test architectures of different sizes.
In particular, we define 4 categories of models based on
the number of parameters. We call those categories
XS, S, M and L. See \Cref{table:sizes} for the exact numbers.
In this table we also report the \emph{width parameter} $w$
that ensures our architecture has the correct number of parameters.
\input{tab/sizes_new}

\begin{remark} 
    \label{remark:bcop_params}
    For most methods, the number of parameters per layer are about the same.
    There are two exceptions, \BCOP and \Sandwich.
    \BCOP parameterizes the convolution kernel
    with $c$ input channels and $c$ output channels 
    using a matrix of size $c \times c$ and $2(k-1)$ matrices of size $c \times c/2$. Therefore, the number of parameters of a convolution using \BCOP is $kc^2$, less than the $k^2c^2$ parameters of a plain convolution. 
    The \Sandwich layer has about twice as many parameters
    as the other layers for the same width,
    as it parameterizes two weight matrices, $A$ and $B$ in \Cref{eq:sandwich},
    per layer.
\end{remark}

\subsection{Loss function}
We use the loss function proposed by \cite{aol}, with the temperature parameter
set to the value used there ($t = 1/4$).
Our goal metric is \cra for perturbation of maximal size $\epsilon = 36/255$.
We aim at robustness of maximal size $2\epsilon$ during training.
In order to achieve that we
set the margin parameter to $2 \sqrt{2} \epsilon$.

\subsection{Optimizer}
We use SGD with a momentum of 0.9 for all experiments. We also used a learning rate schedule. We choose to use \emph{OneCycleLR}, as described by \cite{onecyclelr}, with default values as in \emph{PyTorch}.
We set the batch size to 256 for all experiments.

\subsection{Training Time}
\label{sec:training-time}
On of our main goals is to evaluate what is the best model to use
given a certain time budget.
In order to do this, 
we measure the time per epoch 
as described in \Cref{sec:metrics}
on an A100 GPU with 80GB memory
for different methods and different model sizes. 
Then we estimate the number of
epochs we can do in our chosen time budget of 
either $2$h, $10$h or $24$h,
and use that many epochs to train our models. 
The amount of epochs corresponding to the given time budget 
is summarized in Table~\ref{tab:budget}.
\input{tab/epochs_budget}

\subsection{Hyperparameter Random Search}
\label{sec:greedy-search}
The learning rate and weight decay for each setting 
(model size, method and dataset) 
was tuned on the validation set.
For each method we did hyperparameter search by training
for $2h$ (corresponding number of epochs in \Cref{tab:budget}).
We did 16 runs with learning-rate of the form $10^x$, where $x$ is sampled uniformly in the interval $[-4, -1]$, and with  weight-decay of the form $10^x$, where $x$ is sampled uniformly in the interval $[-5.5, -3.5]$. Finally, we selected the learning rate and weight decay corresponding to the run with the highest validation
\cra for radius $36/255$.
We use these hyperparameters found also for the experiments
with longer training time.


\subsection{Datasets}
We evaluate on three different datasets, 
\CIFARS, \CIFARL \cite{cifar}
and \TinyIN \cite{tiny-imagenet}.

For \CIFARS and \CIFARL we use the architecture 
described in \Cref{table:architecture}.
Since the architectures are identical, 
so are time- and memory requirements, 
and therefore also the epoch budget.
%
As preprocessing we subtract the dataset channel means from each image.
As data augmentation at training time we apply 
random crops (4 pixels) and random flipping. \medskip



In order to assess the behavior on larger images, 
we replicate the evaluation on the \TinyIN dataset \cite{tiny-imagenet}: 
a subset of $200$ classes of the ImageNet \cite{deng2009imagenet} dataset,
with images scaled to have size $64\times64$.

In order to allow for the larger input size of this dataset,
we add one additional \emph{Downsize Block} to our model.
We also divide the width parameter (given in \Cref{table:sizes})
by 2 to keep the amount of parameters similar.
We again subtract the channel mean for each image.
As data augmentation we we us
\emph{RandAugment} \cite{cubuk2020randaugment}
with $2$ transformations of magnitude $9$ (out of $31$).
\subsection{Metrics}
As described in \Cref{sec:metrics}
we evaluate the methods in terms of three main metrics.
The throughput, the memory requirements as well and the 
\cra a model can achieve in a fixed amount of time.

The evaluation of the \emph{throughput}
is performed on an \texttt{NVIDIA A100 80GB PCIe GPU} \footnote{https://www.nvidia.com/content/dam/en-zz/Solutions/Data-Center/a100/pdf/PB-10577-001\_v02.pdf}.
We measure it by averaging the time it takes to process $100$ batches
(including forward pass, backward pass and parameter update),
and use this value to calculate the average number of examples
a model can process per second.
%
In order to estimate the \inference throughput,
we first evaluate and cache all calculations that
do not depend on the input 
(such as power iterations on the weights).
With this we measure the average time of 
the forward pass of $100$ batches,
and calculate the throughput from that value.

%% file: tab/architecture.tex
\setlength{\tabcolsep}{4pt}
\begin{table}[t]
\begin{center}
\caption{
    \textbf{Architecture.}
    It depends on 
    width parameter $w$, 
    kernel size $k$ ($k \in \{1, 3\}$) 
    and the number of classes $c$.
    For details of the \emph{Downsize Block} 
    see \cref{table:convblock}.
}
\label{table:architecture}
\begin{tabular}{rl|l}

\noalign{\smallskip} \hline \noalign{\smallskip}
& Layer name & Output size \\
\noalign{\smallskip} \hline \noalign{\smallskip}

& Input                     & $32 \times 32 \times 3$     \\
& Zero Channel Padding      & $32 \times 32 \times w$     \\
& Conv ($1\times1$ kernel size)  & $32 \times 32 \times w$     \\
& Activation                & $32 \times 32 \times w$     \\

& Downsize Block($k$)            & $16 \times 16 \times 2w$     \\
& Downsize Block($k$)            & $8 \times 8 \times 4w$     \\
& Downsize Block($k$)            & $4 \times 4 \times 8w$     \\
& Downsize Block($k$)            & $2 \times 2 \times 16w$     \\
& Downsize Block($1$)            & $1 \times 1 \times 32w$     \\

& Flatten                   & $32w$    \\
& Linear                    & $32w$     \\
& First Channels($c$)             & $c$       \\

\noalign{\smallskip} \hline

\end{tabular}
\end{center}
\end{table}
\setlength{\tabcolsep}{1.4pt}

%% file: tab/conv_block.tex
\setlength{\tabcolsep}{4pt}
\begin{table}[t]
\begin{center}
\caption{Downsize Block($k$) with input size $s \times s \times t$:
}
\label{table:convblock}
\begin{tabular}{rlll}
\hline\noalign{\smallskip}
& Layer name & Kernel size & Output size \\
\noalign{\smallskip}
\hline
\noalign{\smallskip}

\mrtimes{5}
& Conv              & $k \times k$  & $s \times s \times t$ \\
& Activation        & -             & $s \times s \times t$ \\
& First Channels    & -             & $s \times s \times t/2$ \\
& Pixel Unshuffle   & -             & $s/2 \times s /2\times 2t$ \\

\noalign{\smallskip}
\hline
\end{tabular}
\end{center}
\end{table}
\setlength{\tabcolsep}{1.4pt}

%% file: tab/sizes_new.tex
\setlength{\tabcolsep}{4pt}
\begin{table}[t]
\begin{center}
\caption{
    Number of parameters for different model sizes,
    as well as
    the \emph{width parameter} $w$
    such that the architecture in \cref{table:architecture}
    has the correct size.
}
\label{table:sizes}

\begin{tabular}{lcl}
\noalign{\smallskip} \hline \noalign{\smallskip}
Size &  Parameters (millions) & $w$ \\
\noalign{\smallskip} \hline \noalign{\smallskip}
XS  & $1 < p < 2$           & $16$   \\
S   & $4 < p < 8$           & $32$   \\
M   & $16 < p < 32$         & $64$   \\
L   & $\ \ 64 < p < 128$    & $128$   \\
\noalign{\smallskip} \hline
\end{tabular}

\end{center}
\end{table}
\setlength{\tabcolsep}{1.4pt}

%% file: tab/epochs_budget.tex
\begin{table}[h!]
\centering
\caption{
    Budget of training epochs for different 
    model sizes, layer types and datasets. 
    Batch size and training time are set to be $256$ and $2$h 
    respectively for all the architectures.
}
\label{tab:budget}
\begin{tabular}{l|cccc|cccc}
\toprule
 & \multicolumn{4}{c}{CIFAR} & \multicolumn{4}{c}{TinyImageNet} \\
 & XS & S & M & L & XS & S & M & L \\
\midrule
\textbf{AOL} & 837 & 763 & 367 & 83 & 223 & 213 & 123 & 34 \\
\textbf{BCOP} & 127 & 125 & 94 & 24 & 50 & 50 & 39 & 11 \\
\textbf{CPL} & 836 & 797 & 522 & 194 & 240 & 194 & 148 & 63 \\
\textbf{Cayley} & 356 & 214 & 70 & 17 & 138 & 86 & 30 & 8 \\
\textbf{ECO} & 399 & 387 & 290 & 162 & 142 & 131 & 95 & 54 \\
\textbf{LOT} & 222 & 68 & 11 & - & 83 & 29 & 5 & - \\

\textbf{SLL} & 735 & 703 & 353 & 79 & 242 & 194 & 118 & 32 \\
\textbf{SOC} & 371 & 336 & 201 & 77 & 122 & 87 & 63 & 27 \\

\midrule
\textbf{Param.s (M)}\footnotemark[2] & 1.57 & 6.28 & 25.12 & 100.46 & 1.58 & 6.29 & 25.16 & 100.63 \\
\bottomrule
\end{tabular}
\begin{flushleft}
\footnotemark[2] 
\BCOP has less parameters overall, see \Cref{remark:bcop_params}.
\end{flushleft}
\end{table}

%% file: appendix/9_other.tex

\section{Batch Activation Variance} \label{sec:BatchActVar}

As one (simple to compute) sanity check that the models we train are actually
\ols, we consider the 
\emph{batch activation variance}.
For layers that are \ols, we show below that the batch activation variance
cannot increase from one layer to the next.
This gives us a mechanism  to detect (some) issues with trained models,
including numerical ones, conceptual ones as well as problems in the implementation.
\medskip

To compute the batch activation variance we consider a mini-batch
of inputs, and for this mini-batch we consider the outputs of each
layer.
Denote the outputs of layer $l$ as $a_1^{(l)}, \dots, a_b^{(l)}$, 
where $b$ is the batch size.
Then we set
\begin{align}
    &\mu^{(l)} = \frac{1}{b} \sum_{i=1}^{b} a_i^{(l)} \\
    &\operatorname{BatchVar}_l 
        = \frac{1}{b} \sum_{i=1}^{b} \|a_i^{(l)} - \mu^{(l)}\|_2^2,
\end{align}
where the $l_2$ norm is calculated based on the flattened tensor.
Denote layer $l$ as $f_l$. Then we have that
\begin{align}
    \operatorname{BatchVar}_{l+1}
    &= \frac{1}{b} \sum_{i=1}^{b} \|a_i^{(l+1)} - \mu^{(l+1)}\|_2^2  \\
    &\le \frac{1}{b} \sum_{i=1}^{b} \|f_l(a_i^{(l)}) - f_l(\mu^{(l)})\|_2^2  \\
    &\le \frac{1}{b} \sum_{i=1}^{b} \|a_i^{(l)} - \mu^{(l)}\|_2^2 \\
    &=\operatorname{BatchVar}_{l}.
\end{align}
Here, for the first inequality we use that (by definition) $a_i^{(l+1)} = f_l(a_i^{(l)})$
and that the term $\sum_{i=1}^{n} \|a_i^{(l+1)} - x\|_2^2$ is
minimal for $x = \mu^{(l+1)}$.
The second inequality follows from the \ols property.
The equation above shows that the batch activation variance can not increase
from one layer to the next for \ols layers. Therefore, if we see an increase
in experiments that shows that the layer is not actually \ols.
\medskip

As a further check that the layers are \ols we also apply (convolutional)
power iteration to each linear layer after training.

%% file: appendix/6_further_results.tex
\section{Further Experimental Results}
\label{sec:further_results}
In this section, further experiments --not presented in the main paper--- can be found.

\subsection{Different training time budgets}
In this section we report the experimental results for three
different training budgets: $2h$, $10h$ and $24h$.
See the results in \Cref{table:cr_cifar10_full} (\CIFARS), \Cref{table:cr_cifar100_full} (\CIFARL), 
and \Cref{table:tiny_imagenet-cr} (\TinyIN).
%
Each of those tables also reports the best learning rate and 
weight decay found by the random search for each setting.
\begin{figure}[h!]
    \centering
    \includegraphics[width=\columnwidth]{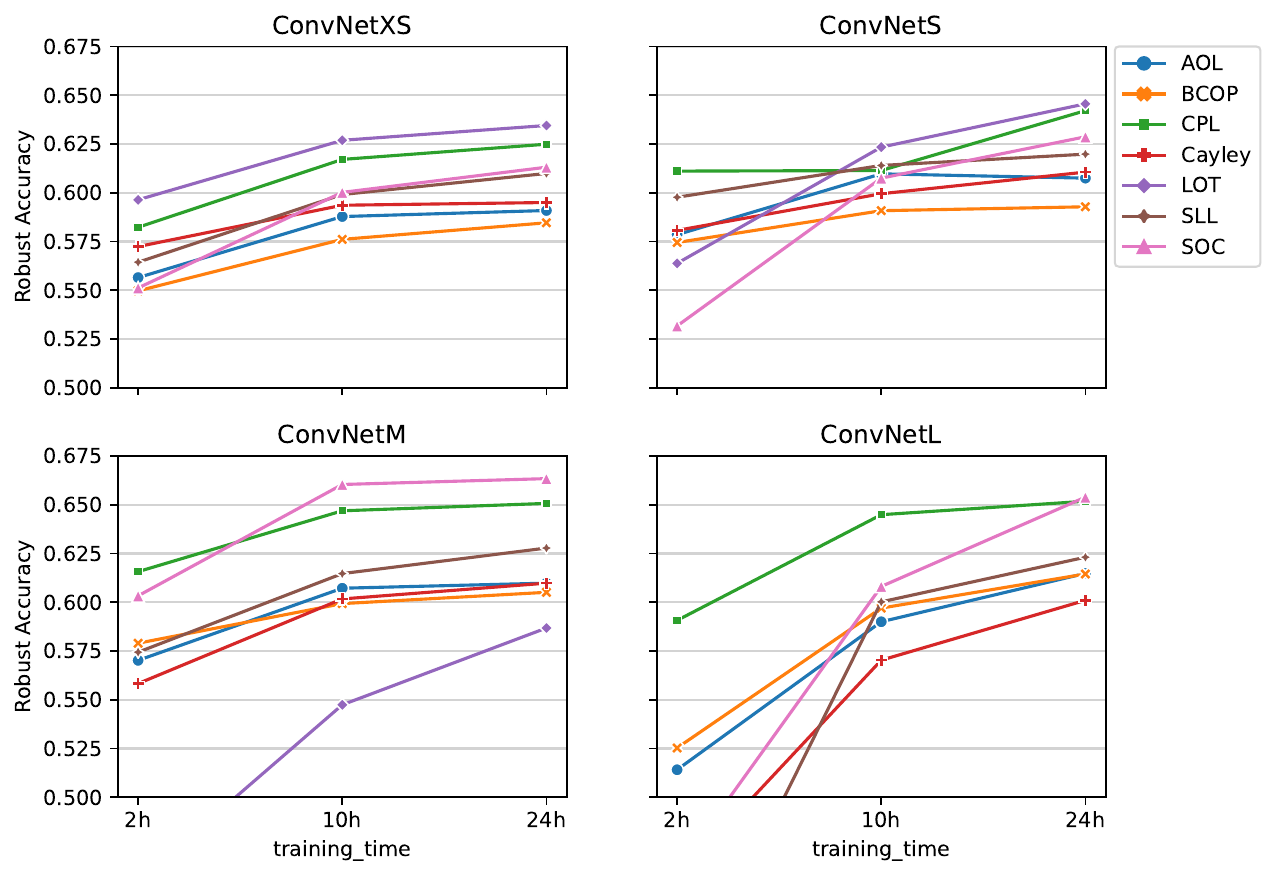}
    \caption{Line plots of the robust accuracy for various methods where models are training with different time budgets
    }
    \label{fig:cr_cifar10}
\end{figure}
Furthermore, a different representation of the impact of the training time on the robust accuracy can be found in \Cref{fig:cr_cifar10}.
\subsection{Time and Memory Requirements on \TinyIN}
See plot \cref{fig:measuring_tiny_imagenet} for an evaluation of time and memory usage for \TinyIN dataset.
The models used on \CIFARS and the ones on \TinyIN 
are identical up to one convolutional block, 
therefore also the results in \Cref{fig:measuring_tiny_imagenet} are
similar to the results on \CIFARS reported in the main paper.

\input{fig/tiny_image_net_times_and_memories}

\subsection{Kernel size $1\times1$} 
For \CIFARS dataset, we tested models where convolutional layers have a kernel size of $1\times1$, to evaluate if the quicker epoch time can compensate for the lower number of parameters.
In almost all cases the answer was negative, 
the version with $3\times3$ kernel outperformed the one with the $1\times1$ kernel.We therefore do not recommend reducing the kernel size.
See \Cref{table:cr_cifar10_full_1x1} for a detailed view of the accuracy and robust accuracy.

\subsection{Clean Accuracy}
\begin{figure*}
\centering
\includegraphics[scale=0.5]{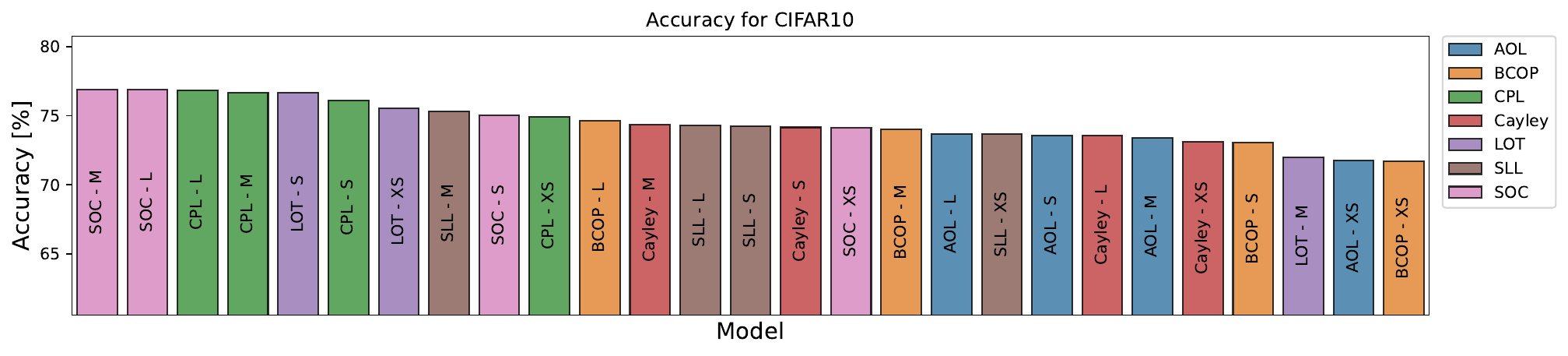}
\includegraphics[scale=0.5, clip, trim=0 0 2.8cm 0]{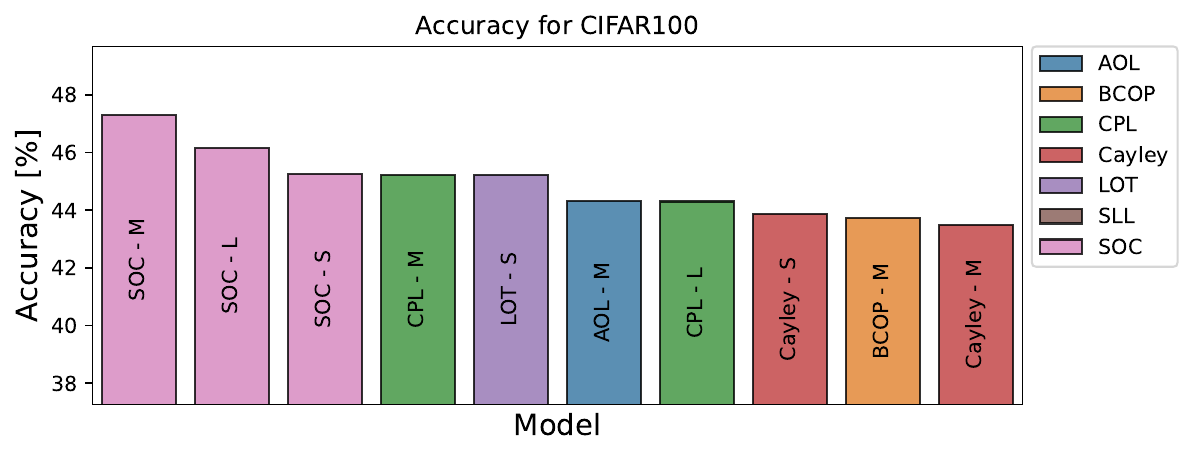}%
\includegraphics[scale=0.5, clip, trim=0 0 2.8cm 0]{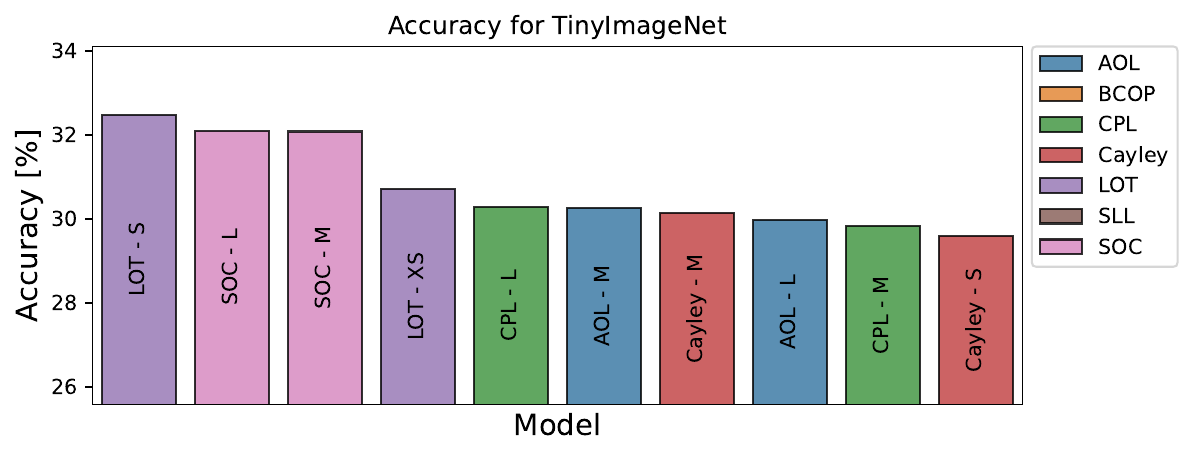}
\caption{
    Barplots of models sorted by decreasing clean accuracy.
    For \CIFARL and \TinyIN only the 10 best performing
    models are shown.}
\label{fig:barplot_accuracy}
\end{figure*}

As a reference, we plotted the accuracy of different models in
\Cref{fig:barplot_accuracy}.
The rankings by accuracy are similar to the rankings by \cra.
One difference is that the \Cayley method performs better relative
to other methods when measured in terms of accuracy.

%% file: fig/tiny_image_net_times_and_memories.tex
\begin{figure}
    \caption{ 
        Measured time and memory requirements on \TinyIN.
    }
    \label{fig:measuring_tiny_imagenet}

    \raggedleft

    \includegraphics[width=0.48\columnwidth]{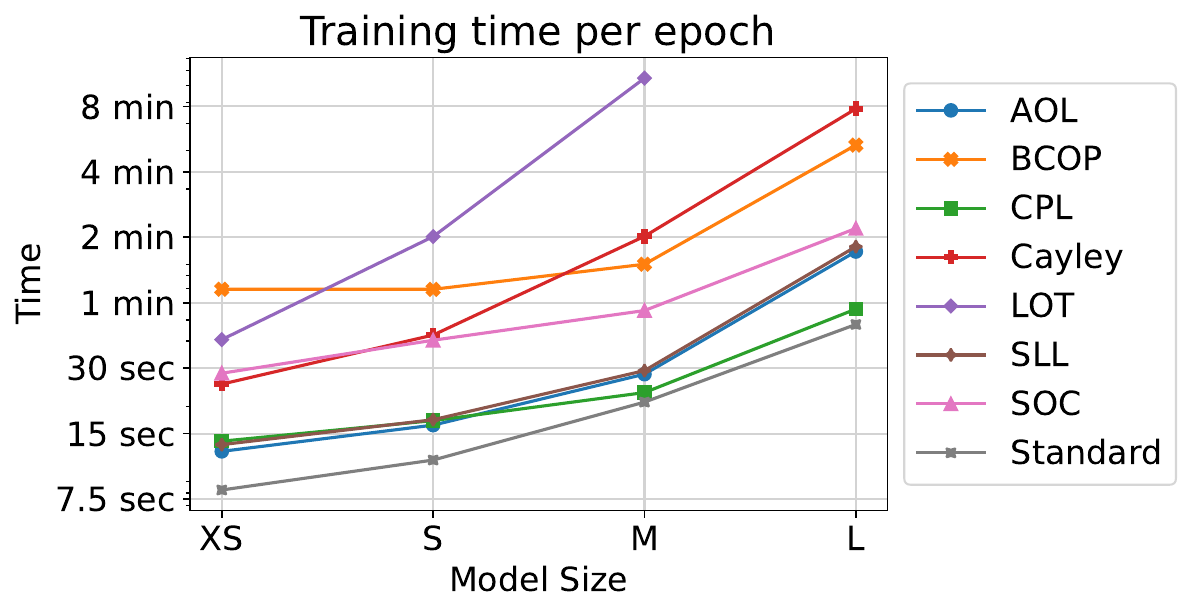}
    \includegraphics[width=0.48\columnwidth]{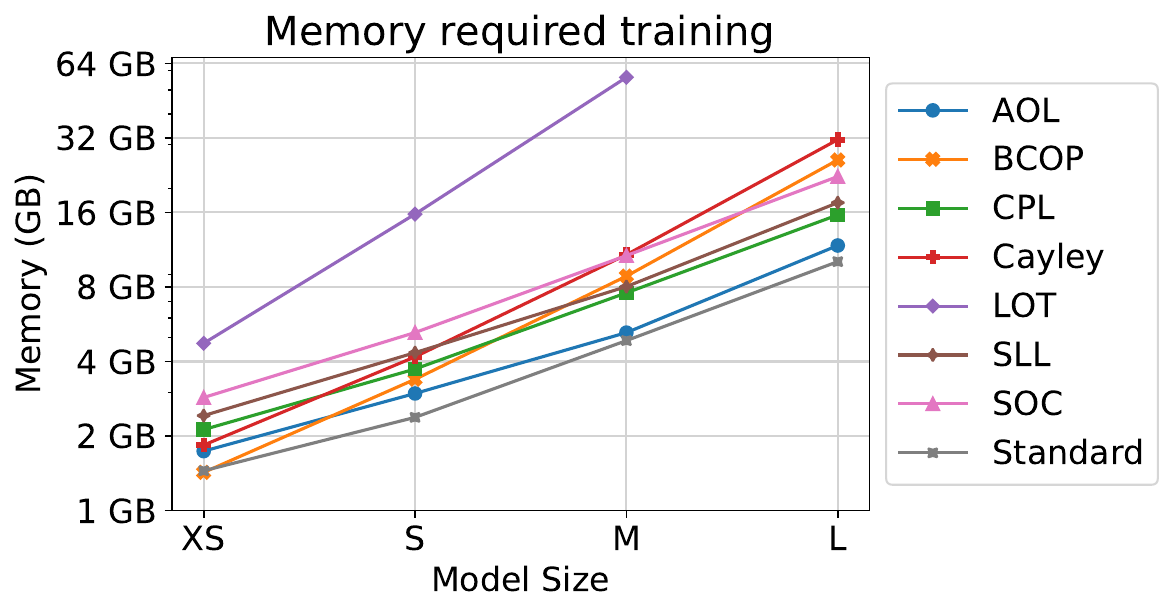}\\
    \hspace{3mm}
    \includegraphics[width=0.45\columnwidth]{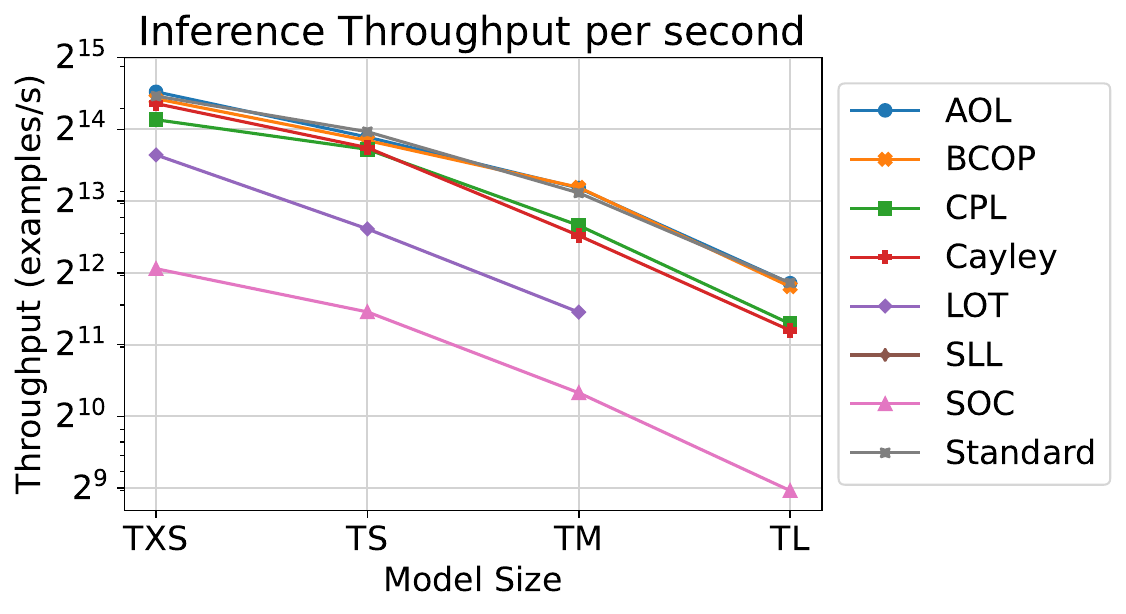}
    \includegraphics[width=0.48\columnwidth]{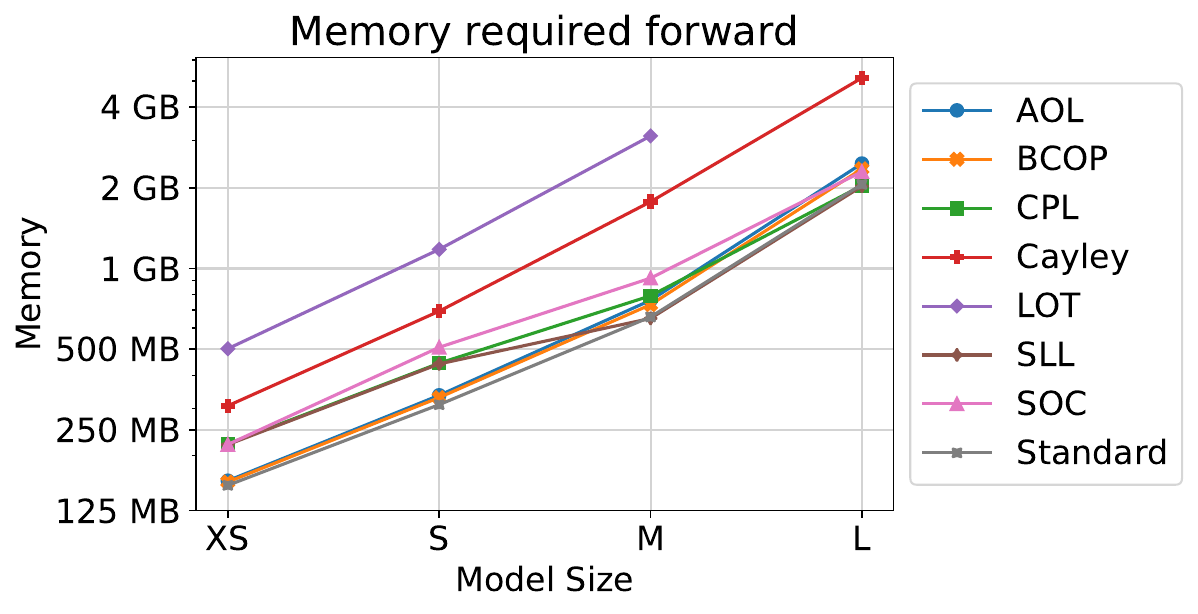}
    
\end{figure}

%% file: appendix/7_reported_issues.tex
\section{Issues and observations}

As already thoroughly analyzed in the dedicated section, some of the known methods in the literature have been omitted in the main paper since we faced serious concerns. Nevertheless, we also encountered some difficulties during the implementation of the methods that we did report in the main paper. The aim of this section is to highlight these difficulties, we hope this can open a constructive debate.

\begin{enumerate}
\item In the \SLL \cite{SLL} code, taken from the authors' repository,  there is no attention to numerical errors that can easily happen from close-to-zero divisions during the parameter transformation. We solve the issue, by adopting the commonly used strategy, i.e. we included a factor of 
\numprint{1e-6}
while dividing for the AOL-rescaling of the weight matrix. Furthermore, the code provided in the \SLL repository only works for a kernel size of $3$, we fixed the issue in our implementation.

\item The \CPL method \cite{CPL} features a high sensitivity to the initialization of the weights. Long training, e.g. 24-hour training, can sometimes result in \emph{NaN} during the update of the weights. In that case we re-ran the models with different seeds.

\item Furthermore, there was no initialization method stated
in the \CPL paper, and also no code was provided. Therefore,
we used the initialization from the similar \SLL method.

\item During the training of \Sandwich we faced some numerical errors. To investigate such errors, we tested a lighter version of the method --- without the learnable rescaling $\Psi$ --- for the reason described in \Cref{obs:sandwich}, which shows that the rescaling $\Psi$ inside the layer can be embedded into the bias term and hence the product $\Psi\Psi^{-1}$ can be omitted. 

\item Similarly, for \SLL, the matrix $Q$ in \Cref{eq:sll} does not add additional 
degrees of freedom to the model. 
Instead of having parameters $W$, $Q$ and $b$ we could define and optimize over
$P = WQ^{-1}$ and $\tilde{b} = Q^{-1}b$.
However, for our experiments we used the original parameterization.

\item The method \Cayley \cite{cayley}, in the form proposed in the original paper, does not cache the --- costly --- transformation of the weight matrix whenever the layer is in inference mode. We fix this issue in our implementation.

\item The \LOT method, \cite{lot}, leverages $10$ inner iterations for both training and inference in order to estimate the inverse of the square root with their proposed Newton-like method. Since the gradient is tracked for the whole procedure, the amount of memory required during the training is prohibitive for large models.
Furthermore, since the memory is required for the parameter
transformation, reducing the batch size does not solve this
problem.
In order to make the Large model (L) fit in the memory, we tested the \LOT method with only $2$ inner iterations. However, the performance in terms of accuracy and robust accuracy is not comparable to other strategies, hence we omitted it from our tables.
\end{enumerate}

\begin{remark}
\label{obs:sandwich}
The learnable parameter $\Psi$ of the sandwich layer corresponds to a scaling of the bias. In details, for each parameters $A,B,b$ and $\Psi=\diag{e^{d_i}}$ there exists a rescaling of the bias $\tilde b$ such that
\begin{equation}
    l(x) = \sqrt{2}A^T\Psi\ReLU{\sqrt{2}\Psi^{-1}Bx +b} = \sqrt{2}A^T\ReLU{\sqrt{2}Bx+\tilde b}
\end{equation}
\begin{proof}
Observing that for each $\alpha>0$ and $x\in\R$, $\ReLU{\alpha x} =\alpha \ReLU{x}$, and that \[
\forall\xx\in\R^n,\quad
\Psi^{-1} \xx = 
\begin{bmatrix}
e^{-d_1}x_1\\
\vdots\\
e^{-d_n}x_n
\end{bmatrix},
\] the following identity holds
\begin{equation}
\begin{aligned}
    l(x) &= \sqrt{2}A^T\Psi\ReLU{\sqrt{2}\Psi^{-1}Bx +b}\\
    &= \sqrt{2}A^T\Psi\ReLU{\sqrt{2}\Psi^{-1}\left(\sqrt{2}Bx +\Psi b\right)}\\
    &= \sqrt{2}A^T\Psi\Psi^{-1}\ReLU{\sqrt{2}Bx +\Psi b}\\
    &= \sqrt{2}A^T\ReLU{\sqrt{2}Bx +\Psi b}.
\end{aligned}
\end{equation}
Considering $\tilde b = \Psi b$ concludes the proof.
\end{proof}
\end{remark}

%% file: appendix/8_code.tex
\section{Code}
We often build on code provided with the original papers.
This includes
\begin{itemize}
    \item \url{https://github.com/berndprach/AOL} (\AOL)
    \item \url{https://github.com/ColinQiyangLi/LConvNet} (\BCOP)
    \item \url{https://github.com/locuslab/orthogonal-convolutions} (\Cayley)
    \item \url{https://github.com/AI-secure/Layerwise-Orthogonal-Training} (\LOT)
    \item \url{https://github.com/acfr/LBDN} (\Sandwich)
    \item \url{https://github.com/araujoalexandre/Lipschitz-SLL-Networks} (\SLL)
    \item \url{https://github.com/singlasahil14/SOC} (\SOC)
\end{itemize}
We are grateful for authors providing code to the research community.

%% file: tab/cr_table_cifar10_full.tex
\begin{table}[t]
\setlength{\tabcolsep}{4pt}
\centering
\caption{
    Standard and Robust Accuracy on \textbf{\CIFARS}. 
    We also report the best learning rate and weight decay
    found by a random search.
}
\label{table:cr_cifar10_full}
\input{tab/raws/CR_table_CIFAR10_full}
\setlength{\tabcolsep}{1.4pt}
\end{table}

%% file: tab/raws/CR_table_CIFAR10_full.tex
\begin{tabular}{llcc|ccc|ccc}
\toprule
 &  &  &  & \multicolumn{3}{c|}{Accuracy [\%]} & \multicolumn{3}{c}{Robust Accuracy [\%]} \\
 &  &  & Training Time & 2h & 10h & 24h & 2h & 10h & 24h \\
Layer & Model & LR & WD &  &  &  &  &  &  \\
\midrule
\textbf{AOL} & \textbf{XS} & \bfseries{\numprint{6e-2}} & \bfseries{\numprint{4e-5}} & 68.5 & 71.7 & 71.7 & 55.6 & 58.8 & \textbf{59.1} \\
\textbf{} & \textbf{S} & \bfseries{\numprint{1e-2}} & \bfseries{\numprint{5e-5}} & 70.9 & 73.0 & 73.6 & 57.9 & \textbf{61.0} & 60.8 \\
\textbf{} & \textbf{M} & \bfseries{\numprint{2e-2}} & \bfseries{\numprint{1e-4}} & 70.3 & 73.4 & 73.4 & 57.0 & 60.7 & \textbf{61.0} \\
\textbf{} & \textbf{L} & \bfseries{\numprint{2e-2}} & \bfseries{\numprint{7e-5}} & 64.7 & 71.8 & 73.7 & 51.4 & 59.0 & \textbf{61.5} \\
\textbf{BCOP} & \textbf{XS} & \bfseries{\numprint{7e-3}} & \bfseries{\numprint{3e-4}} & 69.0 & 70.8 & 71.7 & 55.0 & 57.6 & \textbf{58.5} \\
\textbf{} & \textbf{S} & \bfseries{\numprint{4e-3}} & \bfseries{\numprint{8e-5}} & 70.6 & 72.5 & 73.1 & 57.5 & 59.1 & \textbf{59.3} \\
\textbf{} & \textbf{M} & \bfseries{\numprint{6e-3}} & \bfseries{\numprint{7e-6}} & 71.8 & 73.6 & 74.0 & 57.9 & 59.9 & \textbf{60.5} \\
\textbf{} & \textbf{L} & \bfseries{\numprint{1e-3}} & \bfseries{\numprint{9e-6}} & 67.6 & 73.2 & 74.6 & 52.5 & 59.7 & \textbf{61.5} \\
\textbf{CPL} & \textbf{XS} & \bfseries{\numprint{3e-2}} & \bfseries{\numprint{1e-4}} & 71.7 & 74.3 & 74.9 & 58.2 & 61.7 & \textbf{62.5} \\
\textbf{} & \textbf{S} & \bfseries{\numprint{7e-2}} & \bfseries{\numprint{1e-4}} & 73.9 & 74.1 & 76.1 & 61.1 & 61.1 & \textbf{64.2} \\
\textbf{} & \textbf{M} & \bfseries{\numprint{8e-2}} & \bfseries{\numprint{1e-4}} & 74.3 & 76.4 & 76.6 & 61.6 & 64.7 & \textbf{65.1} \\
\textbf{} & \textbf{L} & \bfseries{\numprint{5e-2}} & \bfseries{\numprint{3e-4}} & 72.6 & 76.5 & 76.8 & 59.1 & 64.5 & \textbf{65.2} \\
\textbf{Cayley} & \textbf{XS} & \bfseries{\numprint{2e-2}} & \bfseries{\numprint{2e-5}} & 71.3 & 73.2 & 73.1 & 57.2 & 59.4 & \textbf{59.5} \\
\textbf{} & \textbf{S} & \bfseries{\numprint{1e-2}} & \bfseries{\numprint{1e-5}} & 71.9 & 73.6 & 74.2 & 58.1 & 60.0 & \textbf{61.1} \\
\textbf{} & \textbf{M} & \bfseries{\numprint{1e-2}} & \bfseries{\numprint{6e-6}} & 70.2 & 73.5 & 74.4 & 55.8 & 60.2 & \textbf{61.0} \\
\textbf{} & \textbf{L} & \bfseries{\numprint{8e-3}} & \bfseries{\numprint{2e-4}} & 61.3 & 71.1 & 73.6 & 45.9 & 57.0 & \textbf{60.1} \\
\textbf{LOT} & \textbf{XS} & \bfseries{\numprint{8e-2}} & \bfseries{\numprint{4e-6}} & 73.5 & 75.2 & 75.5 & 59.6 & 62.7 & \textbf{63.4} \\
\textbf{} & \textbf{S} & \bfseries{\numprint{3e-2}} & \bfseries{\numprint{7e-5}} & 70.5 & 75.0 & 76.6 & 56.4 & 62.3 & \textbf{64.6} \\
\textbf{} & \textbf{M} & \bfseries{\numprint{2e-2}} & \bfseries{\numprint{9e-6}} & 61.4 & 69.3 & 72.0 & 45.7 & 54.7 & \textbf{58.7} \\
\textbf{SLL} & \textbf{XS} & \bfseries{\numprint{9e-2}} & \bfseries{\numprint{2e-5}} & 69.9 & 73.1 & 73.7 & 56.4 & 59.9 & \textbf{61.0} \\
\textbf{} & \textbf{S} & \bfseries{\numprint{4e-2}} & \bfseries{\numprint{2e-4}} & 72.9 & 74.0 & 74.2 & 59.8 & 61.4 & \textbf{62.0} \\
\textbf{} & \textbf{M} & \bfseries{\numprint{9e-2}} & \bfseries{\numprint{9e-5}} & 70.5 & 74.4 & 75.3 & 57.4 & 61.5 & \textbf{62.8} \\
\textbf{} & \textbf{L} & \bfseries{\numprint{6e-2}} & \bfseries{\numprint{2e-4}} & 56.7 & 72.7 & 74.3 & 38.9 & 60.0 & \textbf{62.3} \\
\textbf{SOC} & \textbf{XS} & \bfseries{\numprint{5e-2}} & \bfseries{\numprint{7e-6}} & 68.9 & 72.9 & 74.1 & 55.1 & 60.0 & \textbf{61.3} \\
\textbf{} & \textbf{S} & \bfseries{\numprint{2e-2}} & \bfseries{\numprint{2e-5}} & 67.3 & 73.3 & 75.0 & 53.2 & 60.8 & \textbf{62.9} \\
\textbf{} & \textbf{M} & \bfseries{\numprint{7e-2}} & \bfseries{\numprint{1e-4}} & 73.1 & 77.0 & 76.9 & 60.3 & 66.0 & \textbf{66.3} \\
\textbf{} & \textbf{L} & \bfseries{\numprint{2e-2}} & \bfseries{\numprint{2e-4}} & 62.3 & 73.8 & 76.9 & 46.2 & 60.8 & \textbf{65.4} \\
\bottomrule
\end{tabular}

%% file: tab/cr_table_cifar100_full.tex
\begin{table}[b]
\setlength{\tabcolsep}{4pt}
\centering
\caption{
    Standard and Robust Accuracy on \textbf{\CIFARL}. 
    We also report the best learning rate and weight decay
    found by a random search.
}

\label{table:cr_cifar100_full}
\input{tab/raws/CR_table_CIFAR100_full}
\setlength{\tabcolsep}{1.4pt}
\end{table}

%% file: tab/raws/CR_table_CIFAR100_full.tex
\begin{tabular}{llcc|ccc|ccc}
\toprule
 &  &  &  & \multicolumn{3}{c|}{Accuracy [\%]} & \multicolumn{3}{c}{Robust Accuracy [\%]} \\
 &  &  & Training Time & 2h & 10h & 24h & 2h & 10h & 24h \\
Layer & Model & LR & WD &  &  &  &  &  &  \\
\midrule
\textbf{AOL} & \textbf{XS} & \bfseries{\numprint{3e-2}} & \bfseries{\numprint{1e-4}} & 37.9 & 40.1 & 40.3 & 26.6 & \textbf{28.0} & 27.9 \\
\textbf{} & \textbf{S} & \bfseries{\numprint{2e-2}} & \bfseries{\numprint{2e-5}} & 40.5 & 43.4 & 43.4 & 29.0 & 30.8 & \textbf{31.0} \\
\textbf{} & \textbf{M} & \bfseries{\numprint{7e-2}} & \bfseries{\numprint{6e-6}} & 40.5 & 43.5 & 44.3 & 28.4 & 31.1 & \textbf{31.4} \\
\textbf{} & \textbf{L} & \bfseries{\numprint{6e-2}} & \bfseries{\numprint{4e-5}} & 34.5 & 41.1 & 41.9 & 23.1 & 29.2 & \textbf{29.7} \\
\textbf{BCOP} & \textbf{XS} & \bfseries{\numprint{8e-3}} & \bfseries{\numprint{3e-4}} & 35.4 & 40.0 & 41.4 & 22.9 & 27.8 & \textbf{28.4} \\
\textbf{} & \textbf{S} & \bfseries{\numprint{6e-3}} & \bfseries{\numprint{3e-4}} & 37.8 & 42.1 & 42.8 & 24.5 & 29.5 & \textbf{30.1} \\
\textbf{} & \textbf{M} & \bfseries{\numprint{2e-3}} & \bfseries{\numprint{2e-4}} & 37.6 & 43.8 & 43.7 & 24.6 & 30.4 & \textbf{31.2} \\
\textbf{} & \textbf{L} & \bfseries{\numprint{4e-3}} & \bfseries{\numprint{1e-5}} & 29.2 & 40.3 & 42.2 & 17.3 & 27.2 & \textbf{29.2} \\
\textbf{CPL} & \textbf{XS} & \bfseries{\numprint{9e-2}} & \bfseries{\numprint{3e-5}} & 39.7 & 42.0 & 42.3 & 27.9 & 29.8 & \textbf{30.1} \\
\textbf{} & \textbf{S} & \bfseries{\numprint{9e-2}} & \bfseries{\numprint{2e-4}} & 42.1 & 1.0 & 1.0 & \textbf{29.8} & 0.0 & 0.0 \\
\textbf{} & \textbf{M} & \bfseries{\numprint{4e-2}} & \bfseries{\numprint{4e-5}} & 40.5 & 44.5 & 45.2 & 27.4 & 32.4 & \textbf{33.2} \\
\textbf{} & \textbf{L} & \bfseries{\numprint{9e-2}} & \bfseries{\numprint{2e-4}} & 40.1 & 43.9 & 44.3 & 27.3 & 31.5 & \textbf{32.1} \\
\textbf{Cayley} & \textbf{XS} & \bfseries{\numprint{4e-2}} & \bfseries{\numprint{2e-5}} & 41.1 & 41.6 & 42.3 & 27.9 & 29.2 & \textbf{29.2} \\
\textbf{} & \textbf{S} & \bfseries{\numprint{2e-2}} & \bfseries{\numprint{1e-4}} & 42.3 & 43.1 & 43.9 & 28.8 & 30.4 & \textbf{30.5} \\
\textbf{} & \textbf{M} & \bfseries{\numprint{6e-3}} & \bfseries{\numprint{4e-6}} & 38.6 & 43.3 & 43.5 & 25.3 & 29.9 & \textbf{30.5} \\
\textbf{} & \textbf{L} & \bfseries{\numprint{5e-3}} & \bfseries{\numprint{2e-5}} & 26.3 & 40.3 & 42.9 & 14.3 & 27.0 & \textbf{29.5} \\
\textbf{LOT} & \textbf{XS} & \bfseries{\numprint{6e-2}} & \bfseries{\numprint{3e-4}} & 42.7 & 43.8 & 43.5 & 29.4 & \textbf{30.9} & 30.8 \\
\textbf{} & \textbf{S} & \bfseries{\numprint{5e-2}} & \bfseries{\numprint{2e-4}} & 40.3 & 45.2 & 45.2 & 27.2 & 31.8 & \textbf{32.5} \\
\textbf{} & \textbf{M} & \bfseries{\numprint{4e-2}} & \bfseries{\numprint{9e-5}} & 28.4 & 38.9 & 42.8 & 15.5 & 25.9 & \textbf{29.6} \\
\textbf{SLL} & \textbf{XS} & \bfseries{\numprint{5e-2}} & \bfseries{\numprint{6e-5}} & 37.9 & 40.9 & 41.4 & 25.9 & \textbf{29.0} & 28.9 \\
\textbf{} & \textbf{S} & \bfseries{\numprint{1e-1}} & \bfseries{\numprint{7e-5}} & 40.0 & 41.9 & 42.8 & 28.4 & 29.9 & \textbf{30.5} \\
\textbf{} & \textbf{M} & \bfseries{\numprint{7e-2}} & \bfseries{\numprint{2e-4}} & 39.3 & 41.9 & 42.4 & 27.0 & \textbf{30.0} & 29.9 \\
\textbf{} & \textbf{L} & \bfseries{\numprint{9e-2}} & \bfseries{\numprint{8e-5}} & 22.0 & 38.5 & 42.1 & 10.8 & 26.4 & \textbf{29.6} \\
\textbf{SOC} & \textbf{XS} & \bfseries{\numprint{7e-2}} & \bfseries{\numprint{2e-4}} & 40.7 & 43.1 & 43.1 & 27.7 & 29.7 & \textbf{30.6} \\
\textbf{} & \textbf{S} & \bfseries{\numprint{9e-2}} & \bfseries{\numprint{1e-4}} & 42.2 & 44.5 & 45.2 & 29.3 & 31.9 & \textbf{32.6} \\
\textbf{} & \textbf{M} & \bfseries{\numprint{4e-2}} & \bfseries{\numprint{3e-4}} & 41.8 & 46.2 & 47.3 & 28.8 & 33.5 & \textbf{34.9} \\
\textbf{} & \textbf{L} & \bfseries{\numprint{7e-2}} & \bfseries{\numprint{3e-5}} & 34.7 & 43.1 & 46.2 & 21.5 & 30.2 & \textbf{33.5} \\
\bottomrule
\end{tabular}

%% file: tab/cr_table_tinyimagenet_full.tex
\begin{table}[t]
\setlength{\tabcolsep}{4pt}
\centering
\caption{Experimental results on the \textbf{TinyImageNet} dataset.}
\label{table:tiny_imagenet-cr}
\input{tab/raws/CR_table_TinyImageNet_full}
\setlength{\tabcolsep}{1.4pt}
\end{table}

%% file: tab/raws/CR_table_TinyImageNet_full.tex
\begin{tabular}{llcc|ccc|ccc}
\toprule
 &  &  &  & \multicolumn{3}{c|}{Accuracy [\%]} & \multicolumn{3}{c}{Robust Accuracy [\%]} \\
 &  &  & Training Time & 2h & 10h & 24h & 2h & 10h & 24h \\
Layer & Model & LR & WD &  &  &  &  &  &  \\
\midrule
\textbf{AOL} & \textbf{XS} & \bfseries{\numprint{3e-2}} & \bfseries{\numprint{2e-5}} & 24.5 & 26.9 & 26.6 & 16.5 & \textbf{18.4} & 18.1 \\
\textbf{} & \textbf{S} & \bfseries{\numprint{7e-2}} & \bfseries{\numprint{5e-6}} & 24.9 & 27.3 & 29.3 & 16.8 & 18.5 & \textbf{19.7} \\
\textbf{} & \textbf{M} & \bfseries{\numprint{4e-2}} & \bfseries{\numprint{8e-6}} & 26.2 & 29.5 & 30.3 & 18.1 & 20.4 & \textbf{21.0} \\
\textbf{} & \textbf{L} & \bfseries{\numprint{2e-2}} & \bfseries{\numprint{8e-6}} & 22.1 & 27.7 & 30.0 & 12.8 & 18.8 & \textbf{20.6} \\
\textbf{BCOP} & \textbf{XS} & \bfseries{\numprint{6e-4}} & \bfseries{\numprint{5e-5}} & 11.7 & 20.4 & 22.4 & 4.7 & 11.6 & \textbf{13.8} \\
\textbf{} & \textbf{S} & \bfseries{\numprint{7e-4}} & \bfseries{\numprint{4e-5}} & 19.0 & 25.3 & 26.2 & 10.1 & 15.7 & \textbf{16.9} \\
\textbf{} & \textbf{M} & \bfseries{\numprint{3e-4}} & \bfseries{\numprint{1e-4}} & 20.0 & 25.4 & 27.6 & 10.5 & 15.8 & \textbf{17.2} \\
\textbf{} & \textbf{L} & \bfseries{\numprint{1e-4}} & \bfseries{\numprint{3e-4}} & 9.6 & 23.6 & 27.0 & 2.6 & 14.0 & \textbf{16.8} \\
\textbf{CPL} & \textbf{XS} & \bfseries{\numprint{1e-1}} & \bfseries{\numprint{9e-6}} & 26.5 & 27.5 & 28.3 & 16.3 & 17.5 & \textbf{18.9} \\
\textbf{} & \textbf{S} & \bfseries{\numprint{4e-2}} & \bfseries{\numprint{3e-5}} & 26.2 & 29.3 & 29.3 & 16.7 & 19.4 & \textbf{19.7} \\
\textbf{} & \textbf{M} & \bfseries{\numprint{4e-2}} & \bfseries{\numprint{2e-5}} & 26.3 & 29.4 & 29.8 & 16.7 & 19.5 & \textbf{20.3} \\
\textbf{} & \textbf{L} & \bfseries{\numprint{6e-2}} & \bfseries{\numprint{1e-5}} & 24.7 & 29.8 & 30.3 & 15.0 & 19.2 & \textbf{20.1} \\
\textbf{Cayley} & \textbf{XS} & \bfseries{\numprint{8e-3}} & \bfseries{\numprint{8e-5}} & 25.3 & 27.5 & 27.8 & 15.7 & \textbf{17.9} & \textbf{17.9} \\
\textbf{} & \textbf{S} & \bfseries{\numprint{5e-3}} & \bfseries{\numprint{7e-5}} & 25.6 & 29.3 & 29.6 & 15.8 & 19.1 & \textbf{19.5} \\
\textbf{} & \textbf{M} & \bfseries{\numprint{4e-3}} & \bfseries{\numprint{4e-5}} & 20.2 & 29.2 & 30.1 & 9.9 & 18.8 & \textbf{19.3} \\
\textbf{} & \textbf{L} & \bfseries{\numprint{1e-3}} & \bfseries{\numprint{3e-4}} & 5.7 & 22.1 & 27.2 & 0.9 & 11.9 & \textbf{16.7} \\
\textbf{LOT} & \textbf{XS} & \bfseries{\numprint{3e-2}} & \bfseries{\numprint{2e-4}} & 28.1 & 31.1 & 30.7 & 18.2 & 20.4 & \textbf{20.8} \\
\textbf{} & \textbf{S} & \bfseries{\numprint{5e-2}} & \bfseries{\numprint{1e-4}} & 27.1 & 32.0 & 32.5 & 16.3 & 21.6 & \textbf{21.9} \\
\textbf{} & \textbf{M} & \bfseries{\numprint{2e-2}} & \bfseries{\numprint{6e-6}} & 12.6 & 25.2 & 28.8 & 4.6 & 14.5 & \textbf{18.1} \\
\textbf{SLL} & \textbf{XS} & \bfseries{\numprint{7e-2}} & \bfseries{\numprint{3e-4}} & 24.2 & 25.3 & 25.1 & 14.9 & \textbf{16.7} & 16.6 \\
\textbf{} & \textbf{S} & \bfseries{\numprint{4e-2}} & \bfseries{\numprint{2e-4}} & 24.4 & 25.7 & 27.0 & 15.6 & 16.8 & \textbf{18.4} \\
\textbf{} & \textbf{M} & \bfseries{\numprint{5e-2}} & \bfseries{\numprint{5e-5}} & 17.0 & 26.2 & 26.5 & 7.8 & 17.0 & \textbf{17.7} \\
\textbf{} & \textbf{L} & \bfseries{\numprint{7e-2}} & \bfseries{\numprint{1e-4}} & 11.4 & 26.2 & 27.9 & 2.9 & 16.7 & \textbf{18.8} \\
\textbf{SOC} & \textbf{XS} & \bfseries{\numprint{9e-2}} & \bfseries{\numprint{1e-4}} & 25.5 & 28.7 & 28.9 & 15.7 & 18.7 & \textbf{18.9} \\
\textbf{} & \textbf{S} & \bfseries{\numprint{7e-2}} & \bfseries{\numprint{2e-5}} & 23.9 & 28.4 & 28.8 & 14.0 & 18.5 & \textbf{18.8} \\
\textbf{} & \textbf{M} & \bfseries{\numprint{7e-2}} & \bfseries{\numprint{1e-4}} & 25.7 & 31.1 & 32.1 & 15.5 & 20.4 & \textbf{21.2} \\
\textbf{} & \textbf{L} & \bfseries{\numprint{6e-2}} & \bfseries{\numprint{1e-4}} & 20.3 & 30.1 & 32.1 & 10.1 & 19.7 & \textbf{21.1} \\
\bottomrule
\end{tabular}

%% file: tab/cr_table_cifar10_full_1x1.tex
\begin{table}[t]
\setlength{\tabcolsep}{4pt}
\centering
\caption{
    \textbf{Kernel size $1\times 1$}, on \CIFARS, for different time budgets.
    We report accuracy and \cra for radius $\epsilon=36/355$,
    as well as the best learning rate and weight decay found by a random search.
}
\label{table:cr_cifar10_full_1x1}
\input{tab/raws/CR_table_CIFAR10_full_1x1}
\setlength{\tabcolsep}{1.4pt}
\end{table}

%% file: tab/raws/CR_table_CIFAR10_full_1x1.tex
\begin{tabular}{llcc|ccc|ccc}
\toprule
 &  &  &  & \multicolumn{3}{c|}{Accuracy [\%]} & \multicolumn{3}{c}{Robust Accuracy [\%]} \\
 &  &  & Training Time & 2h & 10h & 24h & 2h & 10h & 24h \\
Layer & Model & LR & WD &  &  &  &  &  &  \\
\midrule
\textbf{AOL} & \textbf{XS} & \bfseries{\numprint{8e-2}} & \bfseries{\numprint{7e-6}} & 66.5 & 67.9 & 68.0 & 52.1 & 53.9 & \textbf{54.3} \\
\textbf{} & \textbf{S} & \bfseries{\numprint{5e-2}} & \bfseries{\numprint{4e-5}} & 67.7 & 69.0 & 69.7 & 53.7 & 55.1 & \textbf{55.6} \\
\textbf{} & \textbf{M} & \bfseries{\numprint{2e-2}} & \bfseries{\numprint{1e-4}} & 67.5 & 69.4 & 70.0 & 54.1 & 55.7 & \textbf{56.7} \\
\textbf{} & \textbf{L} & \bfseries{\numprint{3e-2}} & \bfseries{\numprint{8e-5}} & 64.3 & 68.7 & 69.6 & 50.0 & 54.8 & \textbf{56.1} \\
\textbf{BCOP} & \textbf{XS} & \bfseries{\numprint{1e-2}} & \bfseries{\numprint{7e-5}} & 65.3 & 67.5 & 68.6 & 51.3 & 53.4 & \textbf{54.5} \\
\textbf{} & \textbf{S} & \bfseries{\numprint{4e-3}} & \bfseries{\numprint{2e-5}} & 66.9 & 69.1 & 69.5 & 52.6 & 54.3 & \textbf{55.4} \\
\textbf{} & \textbf{M} & \bfseries{\numprint{8e-3}} & \bfseries{\numprint{5e-6}} & 67.3 & 69.8 & 70.3 & 52.8 & 55.5 & \textbf{56.1} \\
\textbf{} & \textbf{L} & \bfseries{\numprint{4e-3}} & \bfseries{\numprint{2e-4}} & 64.0 & 68.6 & 69.6 & 48.9 & 54.3 & \textbf{55.8} \\
\textbf{BnB} & \textbf{XS} & \bfseries{\numprint{8e-3}} & \bfseries{\numprint{1e-4}} & 66.1 & 66.3 & 66.4 & \textbf{52.1} & 51.2 & 51.5 \\
\textbf{} & \textbf{S} & \bfseries{\numprint{2e-2}} & \bfseries{\numprint{2e-5}} & 67.9 & 69.8 & 69.7 & 53.1 & \textbf{55.8} & 55.3 \\
\textbf{} & \textbf{M} & \bfseries{\numprint{2e-2}} & \bfseries{\numprint{4e-6}} & 67.5 & 70.3 & 71.0 & 52.8 & 56.0 & \textbf{56.8} \\
\textbf{} & \textbf{L} & \bfseries{\numprint{3e-2}} & \bfseries{\numprint{2e-4}} & 66.0 & 66.2 & 61.0 & 51.5 & \textbf{51.7} & 45.5 \\
\textbf{CPL} & \textbf{XS} & \bfseries{\numprint{2e-2}} & \bfseries{\numprint{1e-4}} & 69.2 & 71.5 & 72.0 & 55.5 & 58.6 & \textbf{59.1} \\
\textbf{} & \textbf{S} & \bfseries{\numprint{9e-2}} & \bfseries{\numprint{9e-5}} & 70.6 & 71.2 & 10.0 & 57.8 & \textbf{58.2} & 0.0 \\
\textbf{} & \textbf{M} & \bfseries{\numprint{1e-1}} & \bfseries{\numprint{2e-4}} & 69.7 & 10.0 & 10.0 & \textbf{56.1} & 0.0 & 0.0 \\
\textbf{} & \textbf{L} & \bfseries{\numprint{8e-2}} & \bfseries{\numprint{9e-6}} & 67.8 & 72.3 & 73.8 & 54.0 & 59.4 & \textbf{61.1} \\
\textbf{Cayley} & \textbf{XS} & \bfseries{\numprint{1e-2}} & \bfseries{\numprint{2e-5}} & 65.3 & 67.4 & 67.9 & 50.7 & 52.9 & \textbf{53.8} \\
\textbf{} & \textbf{S} & \bfseries{\numprint{1e-2}} & \bfseries{\numprint{4e-6}} & 65.9 & 68.4 & 68.8 & 51.3 & 53.8 & \textbf{54.5} \\
\textbf{} & \textbf{M} & \bfseries{\numprint{2e-2}} & \bfseries{\numprint{2e-5}} & 63.5 & 67.5 & 69.1 & 48.8 & 53.4 & \textbf{55.0} \\
\textbf{} & \textbf{L} & \bfseries{\numprint{2e-2}} & \bfseries{\numprint{8e-5}} & 57.4 & 64.7 & 67.2 & 41.4 & 49.9 & \textbf{52.9} \\
\textbf{LOT} & \textbf{XS} & \bfseries{\numprint{2e-2}} & \bfseries{\numprint{2e-4}} & 65.8 & 68.0 & 69.3 & 51.1 & 53.9 & \textbf{54.9} \\
\textbf{} & \textbf{S} & \bfseries{\numprint{5e-2}} & \bfseries{\numprint{3e-4}} & 60.0 & 68.1 & 68.8 & 44.4 & 54.3 & \textbf{54.5} \\
\textbf{SLL} & \textbf{XS} & \bfseries{\numprint{4e-2}} & \bfseries{\numprint{1e-4}} & 67.9 & 70.2 & 10.0 & 54.8 & \textbf{57.3} & 0.0 \\
\textbf{} & \textbf{S} & \bfseries{\numprint{3e-2}} & \bfseries{\numprint{2e-4}} & 69.1 & 70.4 & 10.0 & 55.9 & \textbf{57.1} & 0.0 \\
\textbf{} & \textbf{M} & \bfseries{\numprint{7e-2}} & \bfseries{\numprint{2e-4}} & 69.0 & 10.0 & 10.0 & \textbf{55.5} & 0.0 & 0.0 \\
\textbf{} & \textbf{L} & \bfseries{\numprint{9e-2}} & \bfseries{\numprint{2e-4}} & 62.9 & 69.5 & 69.8 & 48.7 & 55.6 & \textbf{56.1} \\
\textbf{SOC} & \textbf{XS} & \bfseries{\numprint{7e-2}} & \bfseries{\numprint{2e-5}} & 65.0 & 67.2 & 67.5 & 49.2 & 52.1 & \textbf{52.3} \\
\textbf{} & \textbf{S} & \bfseries{\numprint{4e-2}} & \bfseries{\numprint{6e-6}} & 65.4 & 68.1 & 68.7 & 49.9 & 52.5 & \textbf{54.2} \\
\textbf{} & \textbf{M} & \bfseries{\numprint{3e-2}} & \bfseries{\numprint{7e-5}} & 66.0 & 69.2 & 70.5 & 50.5 & 54.3 & \textbf{55.6} \\
\textbf{} & \textbf{L} & \bfseries{\numprint{4e-2}} & \bfseries{\numprint{9e-5}} & 64.0 & 69.2 & 70.4 & 48.5 & 54.3 & \textbf{55.9} \\
\bottomrule
\end{tabular}